\documentclass{article}

\usepackage[nonatbib, final]{neurips_2020}
\usepackage{amsmath}
\usepackage{amssymb}
\usepackage{bm,mathtools}

\newcommand\fat[1]{\mathbf{#1}}

\newcommand{\csize}{\fat{c}_{\text{size}}}
\newcommand{\ccrop}{\fat{c}_{\text{crop}}}
\newcommand{\car}{\fat{c}_{\text{ar}}}

\def\eqref#1{equation~\ref{#1}}

\def\1{\bm{1}}

\def\rvc{{\mathbf{c}}}

\def\rvn{{\mathbf{n}}}

\def\rvx{{\mathbf{x}}}

\def\vtheta{{\bm{\theta}}}

\def\vs{{\bm{s}}}

\def\mI{{\bm{I}}}

\DeclareMathAlphabet{\mathsfit}{\encodingdefault}{\sfdefault}{m}{sl}
\SetMathAlphabet{\mathsfit}{bold}{\encodingdefault}{\sfdefault}{bx}{n}

\def\gN{{\mathcal{N}}}

\newcommand{\pdata}{p_{\rm{data}}}

\newcommand{\E}{\mathbb{E}}

\newcommand{\R}{\mathbb{R}}

\DeclarePairedDelimiterX{\infdivx}[2]{(}{)}{%
  #1\;\delimsize\|\;#2%
}

\usepackage[utf8]{inputenc} %
\usepackage[T1]{fontenc}    %
\usepackage{booktabs}       %
\usepackage{amsfonts}       %
\usepackage{nicefrac}       %
\usepackage{microtype}      %
\usepackage{xcolor}      
\usepackage{xspace}%
\usepackage[export]{adjustbox}
\usepackage{siunitx} %
\usepackage{hyperref}       %
\usepackage{url}            %

\usepackage{wrapfig}

\usepackage{placeins}
\usepackage{graphbox} %
\usepackage{breqn}
\usepackage{multirow}
\usepackage{mathtools}
\usepackage{bbm}
\usepackage{tabularx}
\usepackage{booktabs}
\usepackage{algorithm}
\usepackage{algpseudocode}
\usepackage{listings}
\usepackage{svg}

\definecolor{codegreen}{rgb}{0,0.6,0}
\definecolor{codegray}{rgb}{0.5,0.5,0.5}
\definecolor{codepurple}{rgb}{0.58,0,0.82}
\definecolor{backcolour}{rgb}{0.95,0.95,0.92}
\definecolor{codered}{rgb}{0.89,0.4,.45}

\lstdefinestyle{mystyle}{
    backgroundcolor=\color{backcolour},   
    commentstyle=\color{codegreen},
    keywordstyle=\color{codered},
    numberstyle=\tiny\color{codegray},
    stringstyle=\color{codepurple},
    basicstyle=\ttfamily\footnotesize,
    breakatwhitespace=false,         
    breaklines=true,                 
    captionpos=b,                    
    keepspaces=true,                 
    numbers=left,                    
    numbersep=5pt,                  
    showspaces=false,                
    showstringspaces=false,
    showtabs=false,                  
    tabsize=2
}

\usepackage{caption}
\usepackage{subcaption}
\usepackage{cleveref}
\crefrangelabelformat{Section}{#3#1#4--#5#2#6}
\Crefname{figure}{Fig.}{Figs.}
\Crefname{table}{Tab.}{Tabs.}
\Crefname{section}{Sec.}{Secs.}
\Crefname{appendix}{App.}{Apps.}
\Crefname{equation}{Eq.}{Eqs.}
\Crefname{algorithm}{Alg.}{Algs.}
\usepackage{graphicx,multirow}
\usepackage[numbers]{natbib}

\usepackage{pifont}%

\title{\emph{SDXL:} Improving Latent Diffusion Models for High-Resolution Image Synthesis}
\author{%
    Dustin Podell \\
    \And 
    Zion English \\
    \And 
    Kyle Lacey \\
    \And 
    Andreas Blattmann \\
    \And 
    Tim Dockhorn \\
    \And 
    Jonas M\"uller \\
    \And 
    Joe Penna \\
    \And 
    Robin Rombach \\
}

\newcommand\modelname[0]{\emph{SDXL}\xspace}
\newcommand\sd[0]{\emph{Stable Diffusion}\xspace}
\newcommand\sdshort[0]{\emph{SD}\xspace}

\providecommand{\imwidth}{}
\providecommand{\impath}[1]{}
\providecommand{\impatha}[1]{}
\providecommand{\impathb}[1]{}
\providecommand{\impathc}[1]{}

\newcommand{\teaser}{
\begin{figure}[htbp] %
     \centering
     \includegraphics[width=0.94\textwidth]{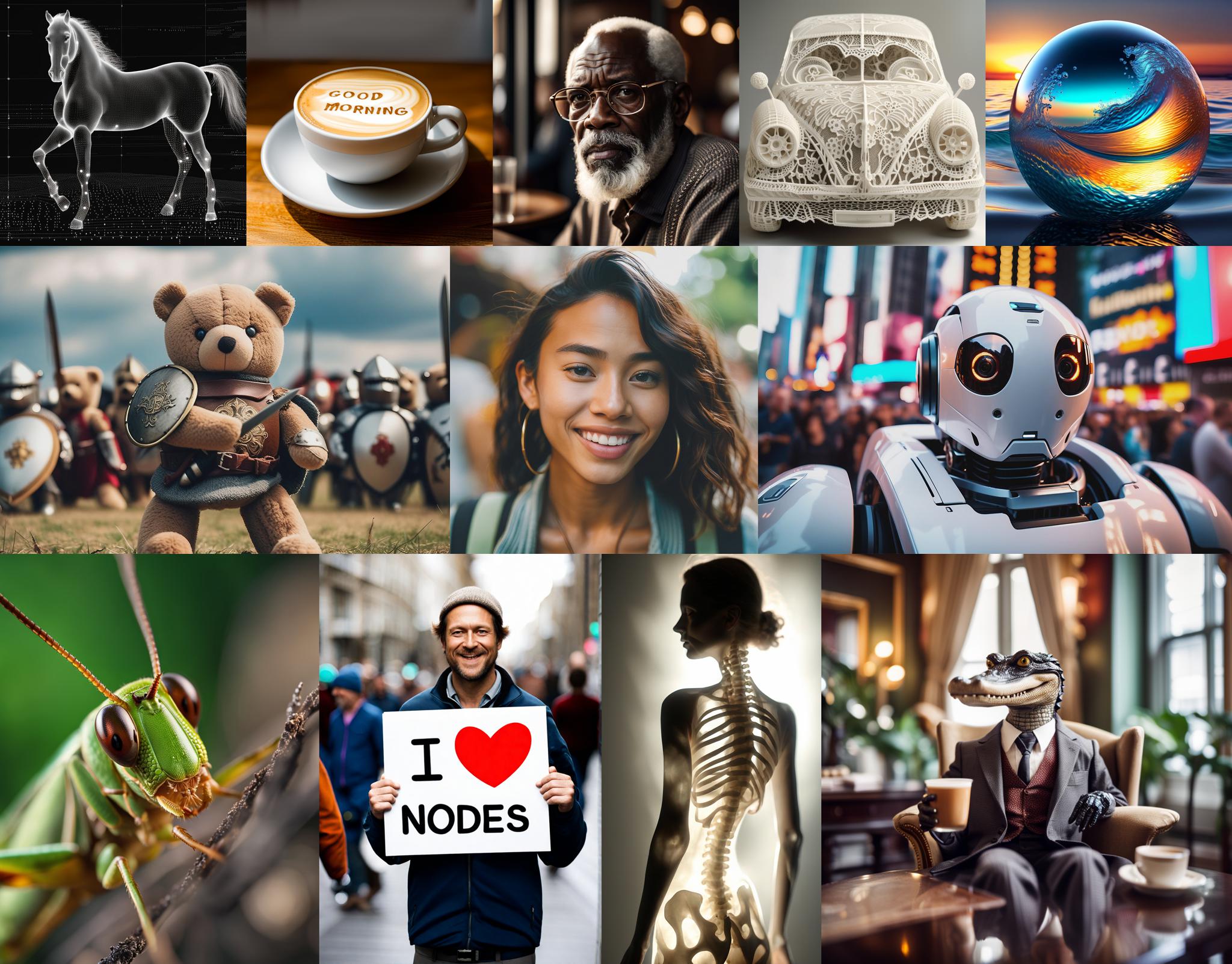}
    \label{fig:teaser}
\end{figure}
}

\newcommand{\sizecondvtwo}{
\begin{figure}[htbp]
\renewcommand{\imwidth}{0.25\textwidth}
\setlength{\tabcolsep}{.5pt}
\centering
\begin{tabular}{cccc}
\toprule
\footnotesize $\csize=(64, 64)$ & \footnotesize $\csize=(128, 128)$, & \footnotesize $\csize=(256,236)$, & \footnotesize $\csize=(512,512)$, \\
\midrule
\includegraphics[width=\imwidth]{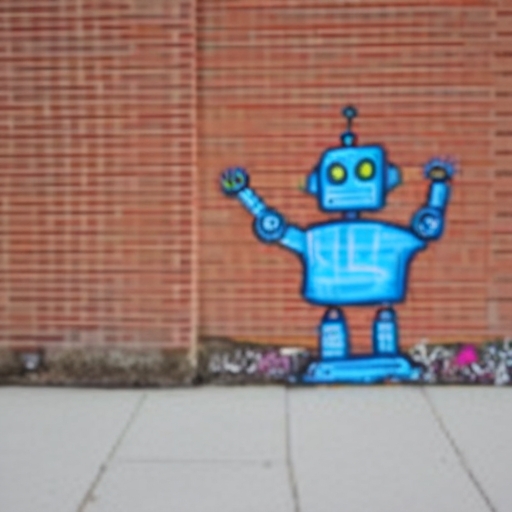} &
\includegraphics[width=\imwidth]{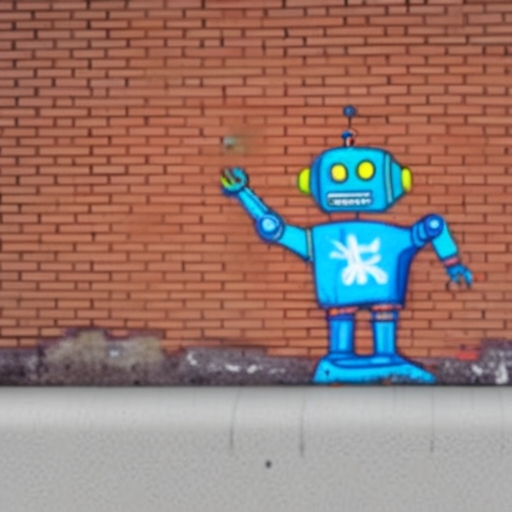} & 
\includegraphics[width=\imwidth]{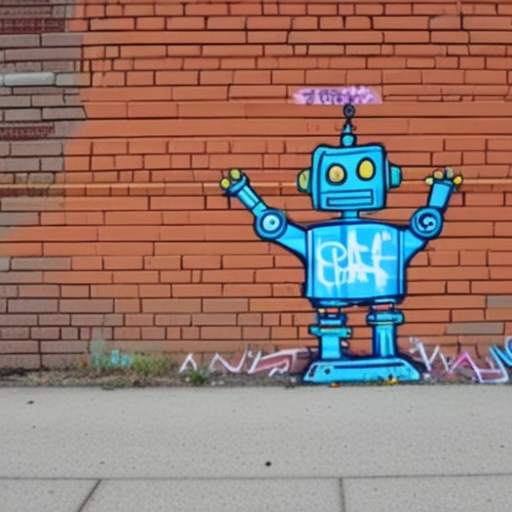} & 
\includegraphics[width=\imwidth]{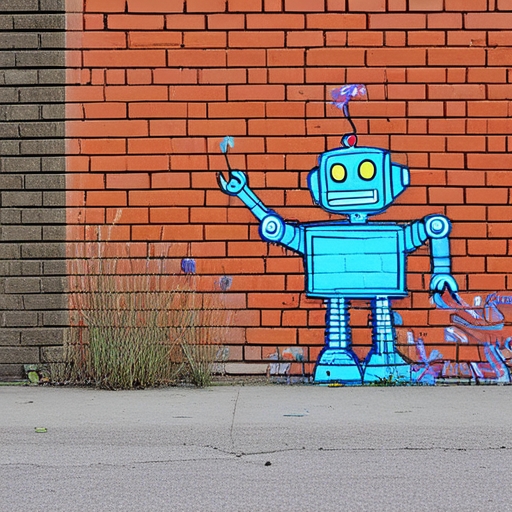} \\
\midrule
\multicolumn{4}{c}{\scriptsize\emph{'A robot painted as graffiti on a brick wall. a sidewalk is in front of the wall, and grass is growing out of cracks in the concrete.'}} \\
\midrule
\includegraphics[width=\imwidth]{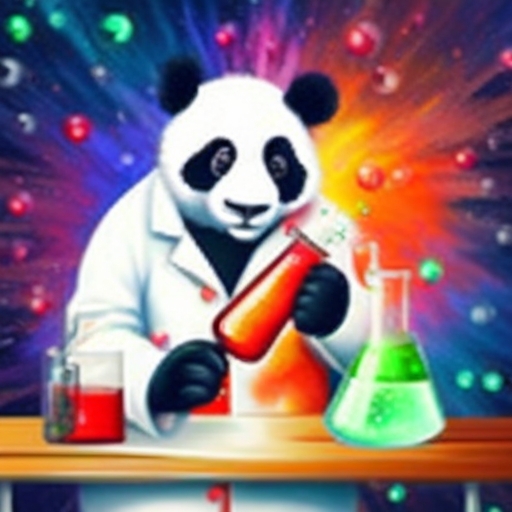} &
\includegraphics[width=\imwidth]{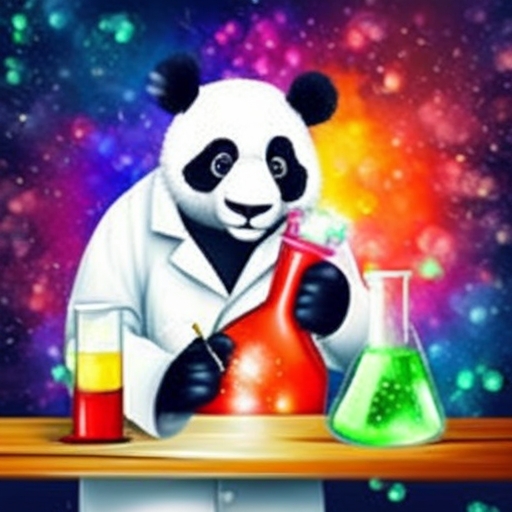} & 
\includegraphics[width=\imwidth]{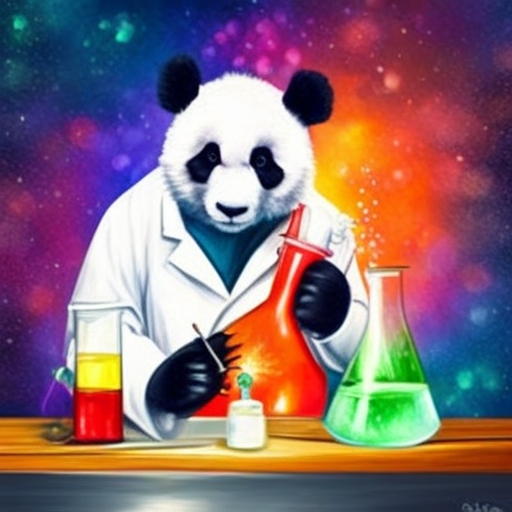} & 
\includegraphics[width=\imwidth]{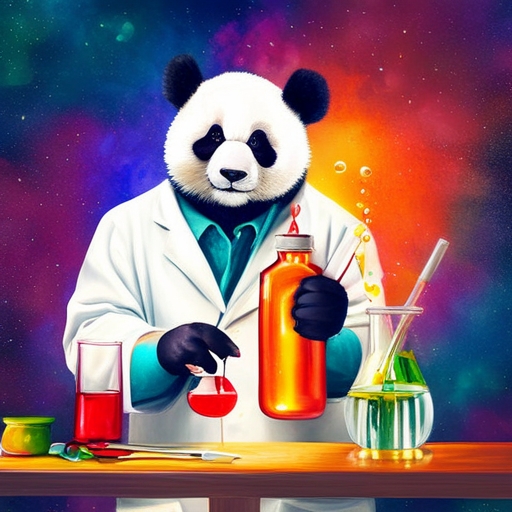} \\
\midrule
\multicolumn{4}{c}{\footnotesize\emph{'Panda mad scientist mixing sparkling chemicals, artstation.'}} \\
\bottomrule
\end{tabular}
\caption{The effects of varying the size-conditioning: We show draw 4 samples with the same random seed from \modelname and vary the size-conditioning as depicted above each column. The image quality clearly increases when conditioning on larger image sizes. Samples from the $512^2$ model, see \Cref{subsec:putting}. Note: For this visualization, we use the $512 \times 512$ pixel base model (see \Cref{subsec:putting}), since the effect of size conditioning is more clearly visible before $1024 \times 1024$ finetuning. Best viewed zoomed in. \vspace{-2em}}
\label{fig:sizecond}
\end{figure}
}

\newcommand{\cropcondvtwo}{
\begin{figure}[htbp]
\renewcommand{\imwidth}{0.25\textwidth}
\setlength{\tabcolsep}{.5pt}
\centering
\begin{tabular}{cccc}
\toprule
\footnotesize $\ccrop=(0,0)$ & \footnotesize $\ccrop=(0,256)$, & \footnotesize $\ccrop=(256,0)$, & \footnotesize $\ccrop=(512,512)$, \\
\midrule
\includegraphics[width=\imwidth]{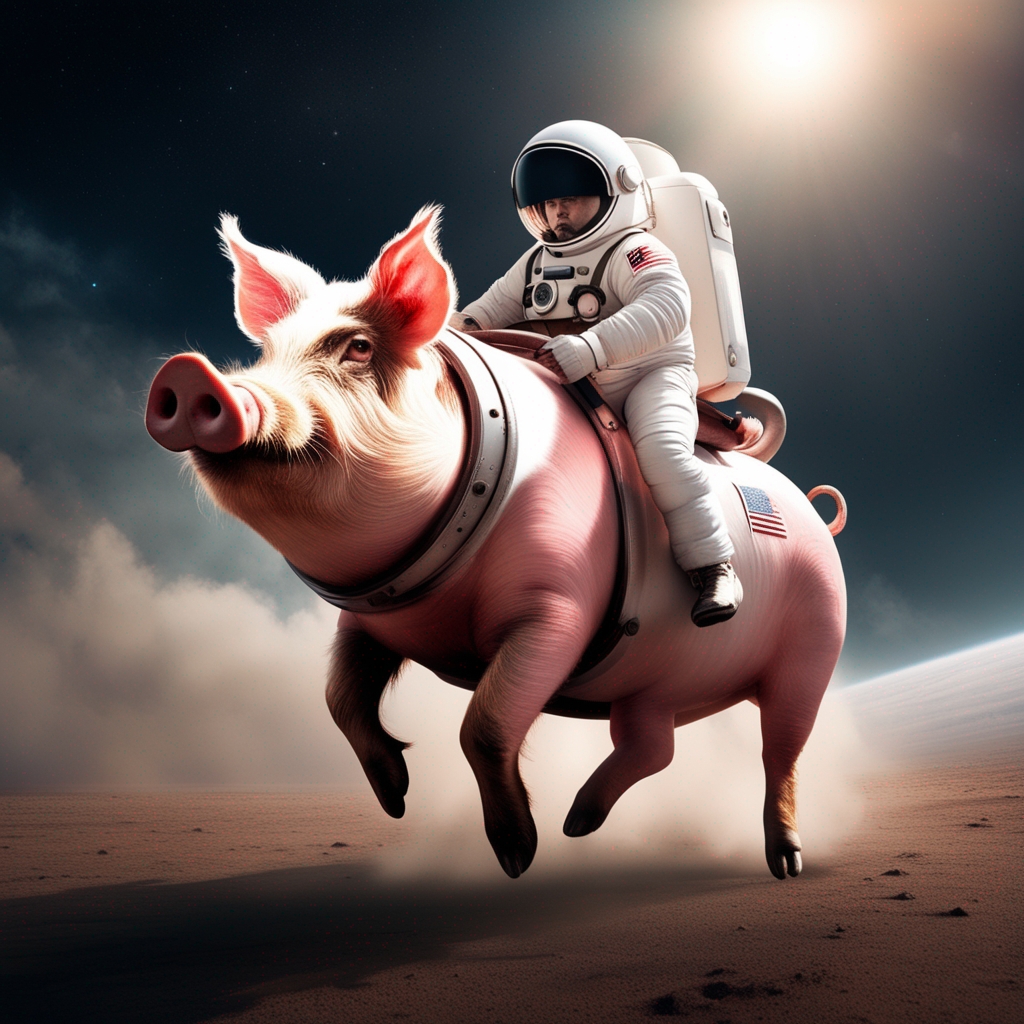} &
\includegraphics[width=\imwidth]{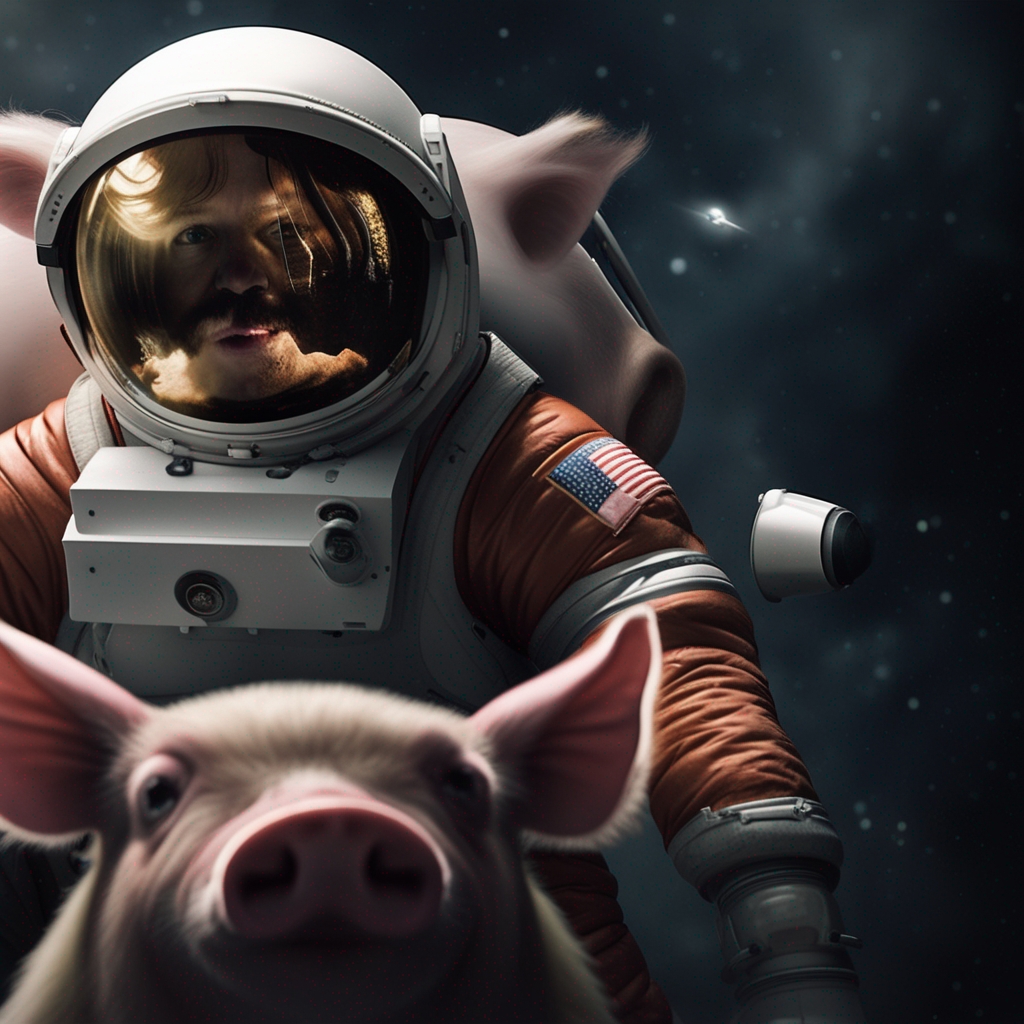} & 
\includegraphics[width=\imwidth]{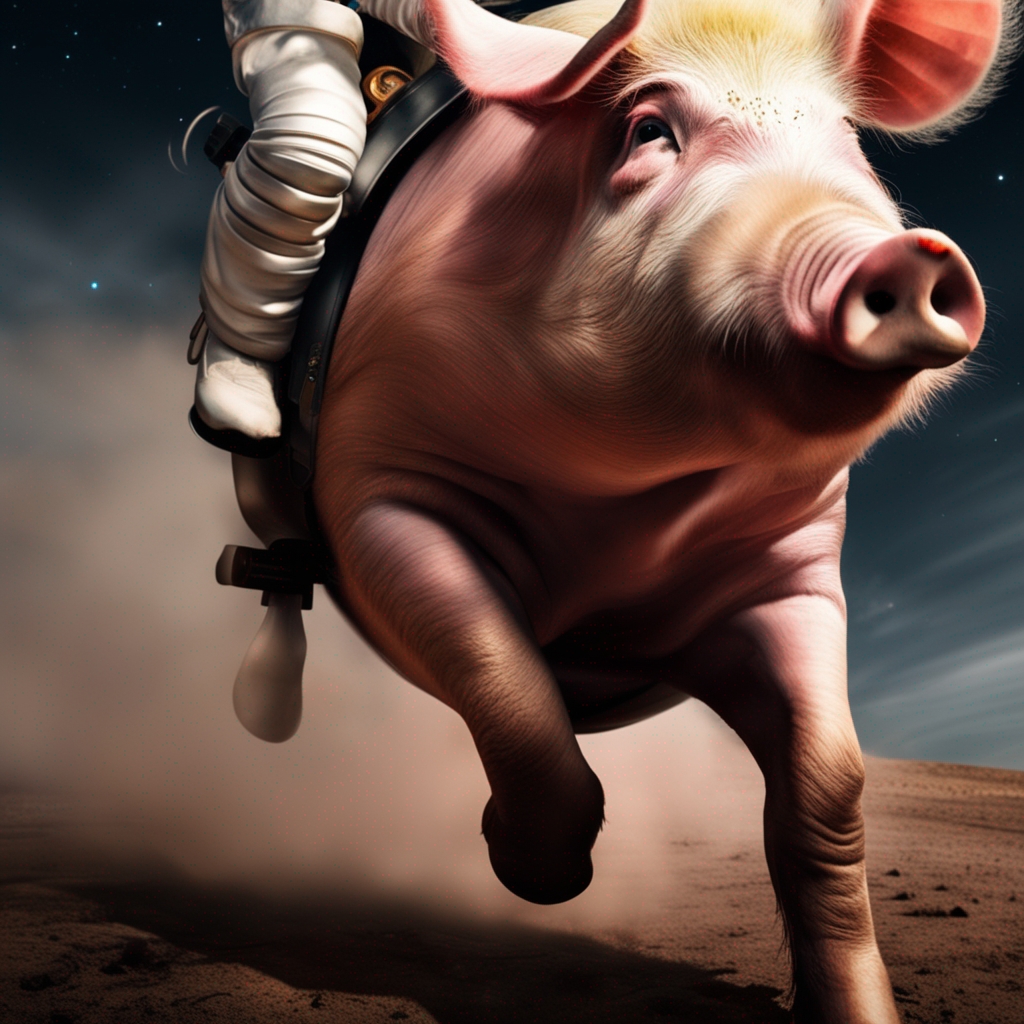} & 
\includegraphics[width=\imwidth]{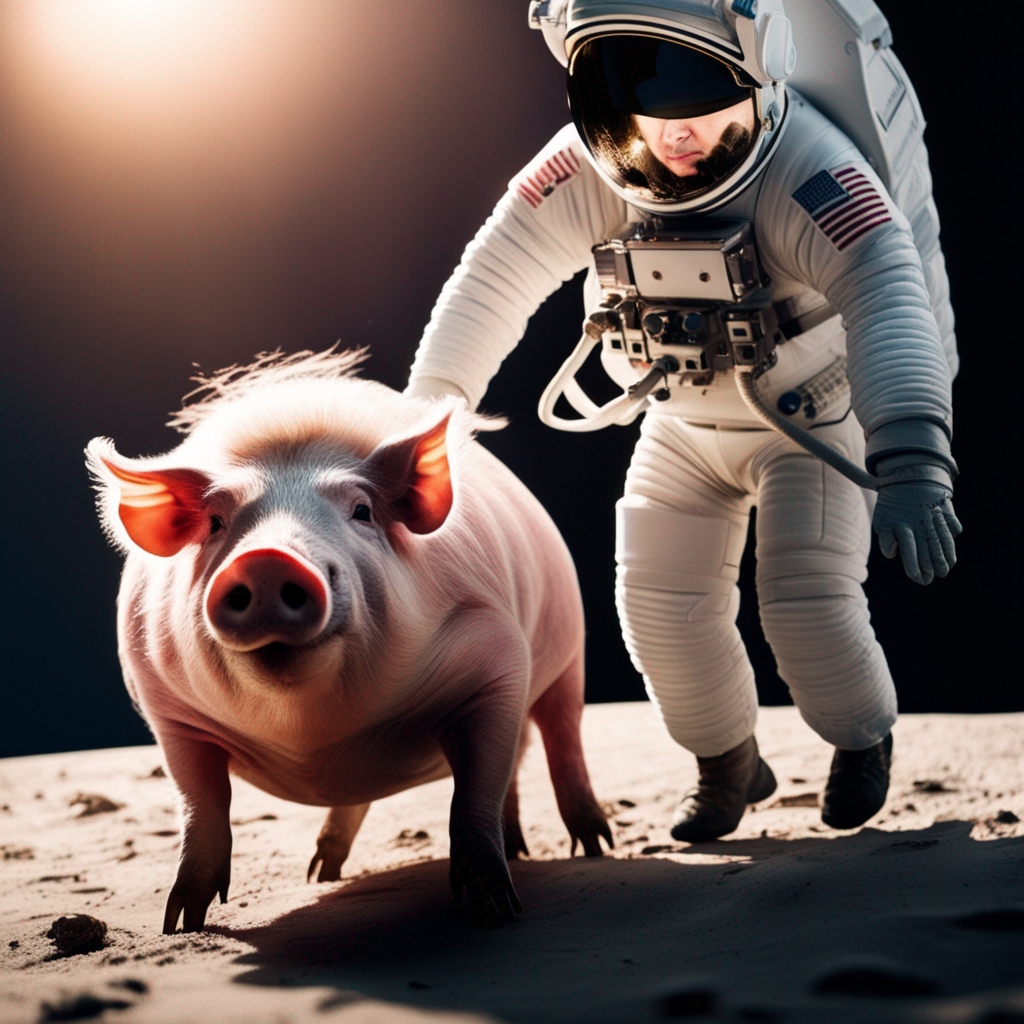} \\
\midrule
\multicolumn{4}{c}{\footnotesize\emph{'An astronaut riding a pig, highly realistic dslr photo, cinematic shot.'}} \\
\midrule
\includegraphics[width=\imwidth]{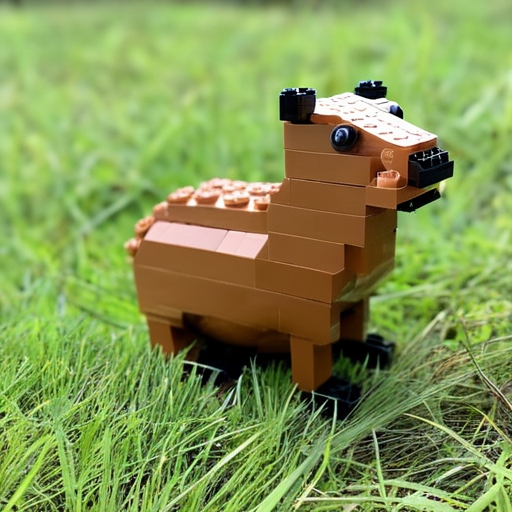} &
\includegraphics[width=\imwidth]{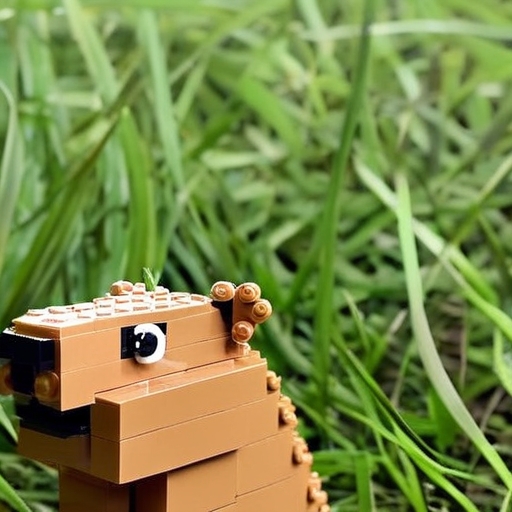} & 
\includegraphics[width=\imwidth]{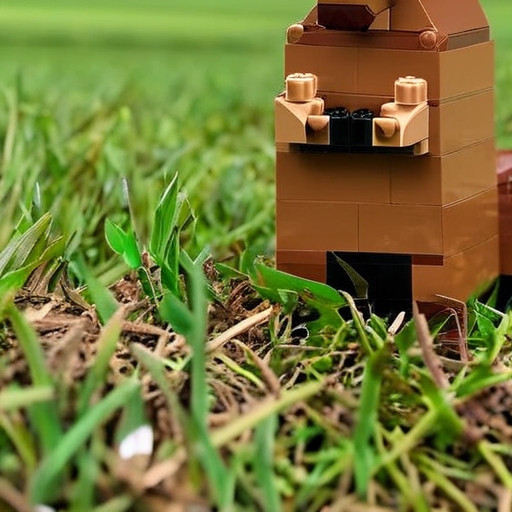} & 
\includegraphics[width=\imwidth]{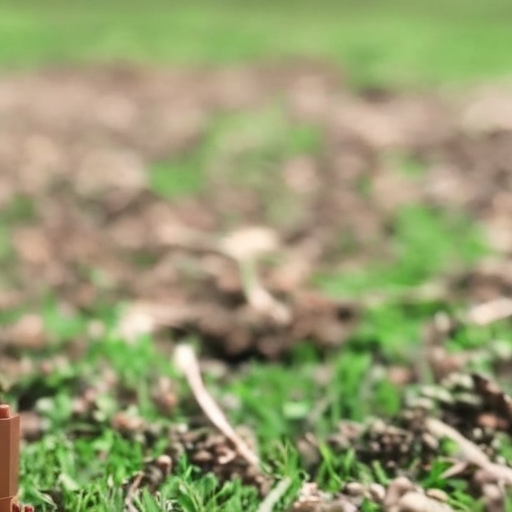} \\
\midrule
\multicolumn{4}{c}{\footnotesize\emph{'A capybara made of lego sitting in a realistic, natural field.'}} \\
\bottomrule
\end{tabular}
\caption{Varying the crop conditioning as discussed in Sec.~\ref{subsec:condtricks}. See \Cref{fig:comp_old_model} and \Cref{fig:comp_old_model_app} for samples from \sdshort 1.5 and \sdshort 2.1 which provide no explicit control of this parameter and thus introduce cropping artifacts. Samples from the $512^2$ model, see \Cref{subsec:putting}.}
\label{fig:cropcond}
\end{figure}
}

\newcommand{\fidvsclip}{
\begin{figure}[htbp]
\centering
\includegraphics[width=.49\textwidth]{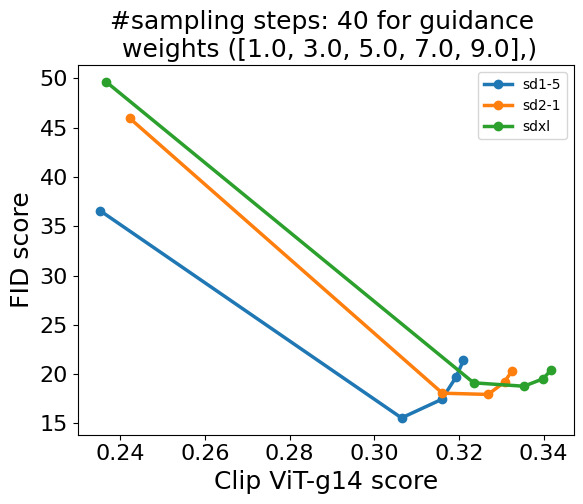}
\hfill
\includegraphics[width=.49\textwidth]{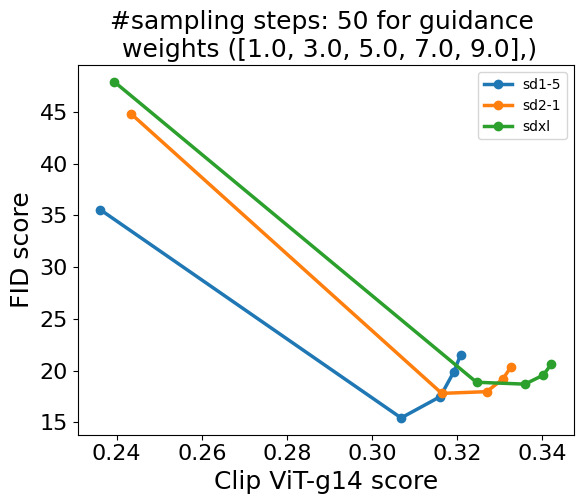}
\caption{\label{fig:fid+vs_clip} Plotting FID vs CLIP score for different cfg scales. \emph{SDXL} shows only slightly improved text-alignment, as measured by CLIP-score, compared to previous versions that do not align with the judgement of human evaluators. Even further and similar as in \cite{kirstain2023pick}, FID are worse than for both \emph{SD-1.5} and \emph{SD-2.1}, while human evaluators clearly prefer the generations of \emph{SD-XL} over those of these previous models.}
\end{figure}
}

\newcommand{\userstudyandmodel}{
\begin{figure}[htbp]
\centering
\includegraphics[align=c,width=.475\textwidth]{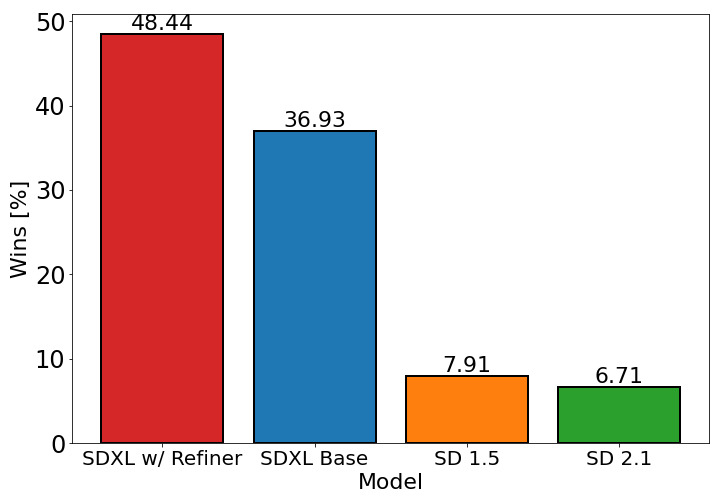}
\includegraphics[align=c,width=.475\textwidth]{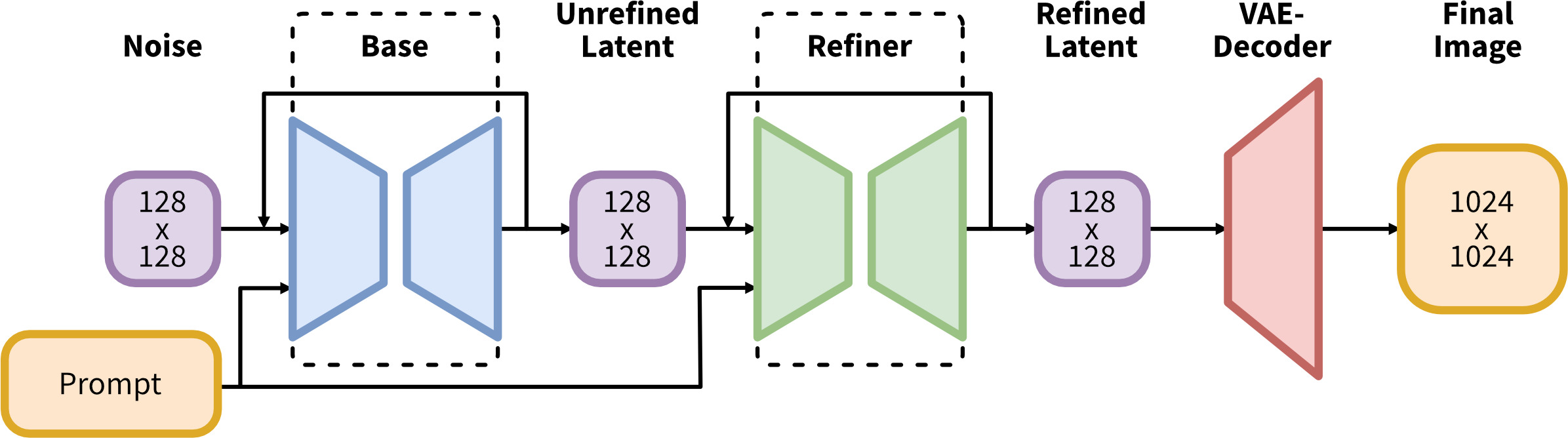}
\caption{\label{fig:userstudyandmodel} \emph{Left:} Comparing user preferences between \modelname and \sd 1.5 \& 2.1. While \modelname already clearly outperforms \sd 1.5 \& 2.1, adding the additional refinement stage boosts performance. \emph{Right:} Visualization of the two-stage pipeline: We generate initial latents of size $128\times128$ using \modelname. Afterwards, we utilize a specialized high-resolution \emph{refinement model} and apply SDEdit~\cite{meng2021sdedit} on the latents generated in the first step, using the same prompt. \modelname and the refinement model use the same autoencoder.}
\end{figure}
}

\newcommand{\sizedist}{
\begin{wrapfigure}{r}{.49\textwidth}
\vspace{-3em}
\renewcommand{\imwidth}{.47\textwidth}
\renewcommand{\impath}[1]{img/##1}
\includegraphics[width=\imwidth]{\impath{size-dist}}
\caption{\label{fig:size_dist} Height-vs-Width distribution of our pre-training dataset. Without the proposed size-conditioning, 39\% of the data would be discarded due to edge lengths smaller than 256 pixels as visualized by the dashed black lines. Color intensity in each visualized cell is proportional to the number of samples.\vspace{-1em}}
\end{wrapfigure}
}

\newcommand{\comparetoif}{
\begin{figure}[htbp]
     \centering
     \includegraphics[height=0.81\textheight,keepaspectratio]{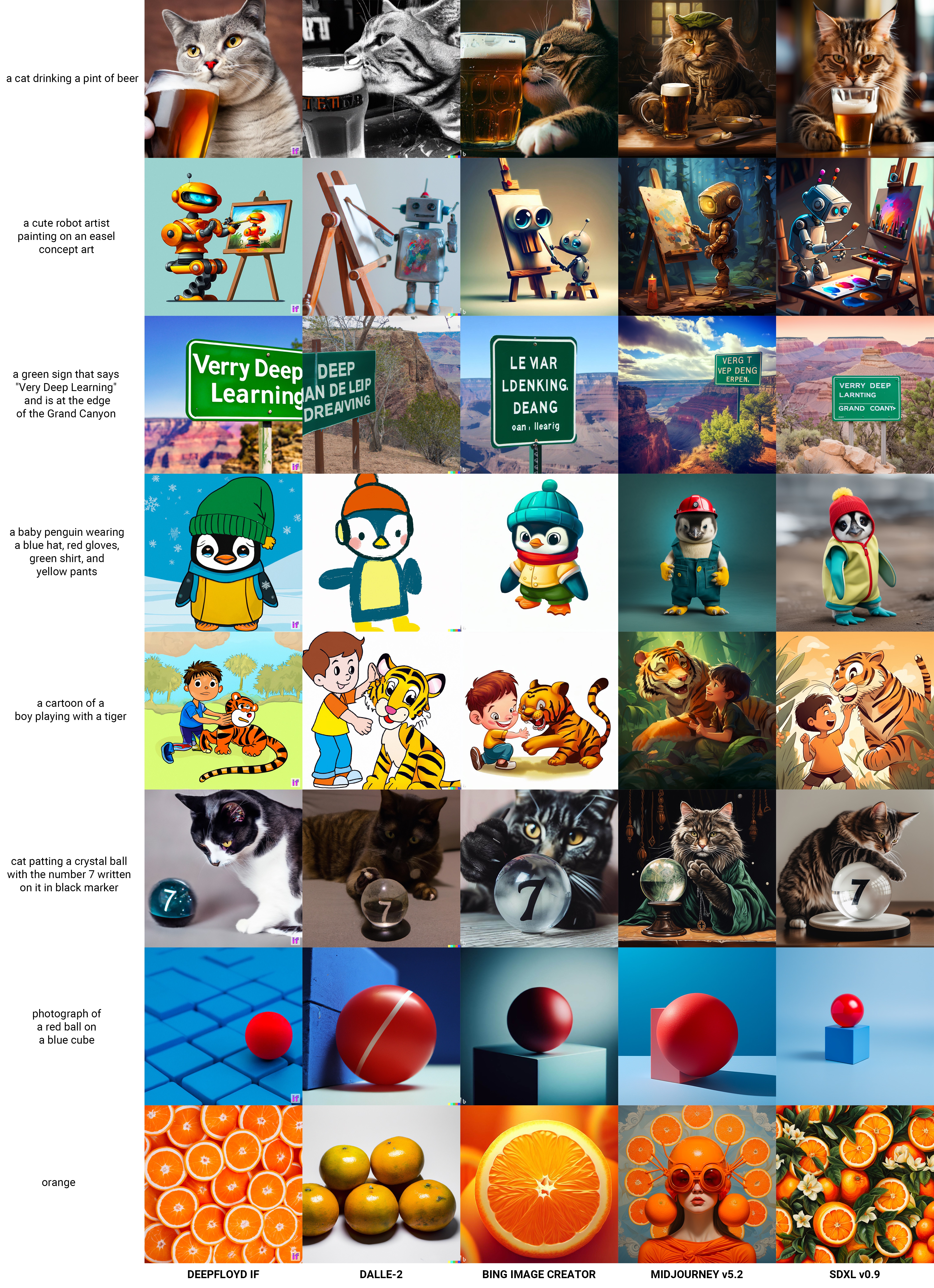}
     \caption{Qualitative comparison of \modelname{} with DeepFloyd IF, DALLE-2, Bing Image Creator, and Midjourney v5.2. To mitigate any bias arising from cherry-picking, Parti (P2) prompts were randomly selected. Seed \num{3} was uniformly applied across all models in which such a parameter could be designated. For models without a seed-setting feature, the first generated image is included.}
     \label{fig:comparetoif}
\end{figure}
}

\newcommand{\refinevsnorefineappendix}{
\begin{figure}[htbp] %
     \centering
     \includegraphics[width=\textwidth]{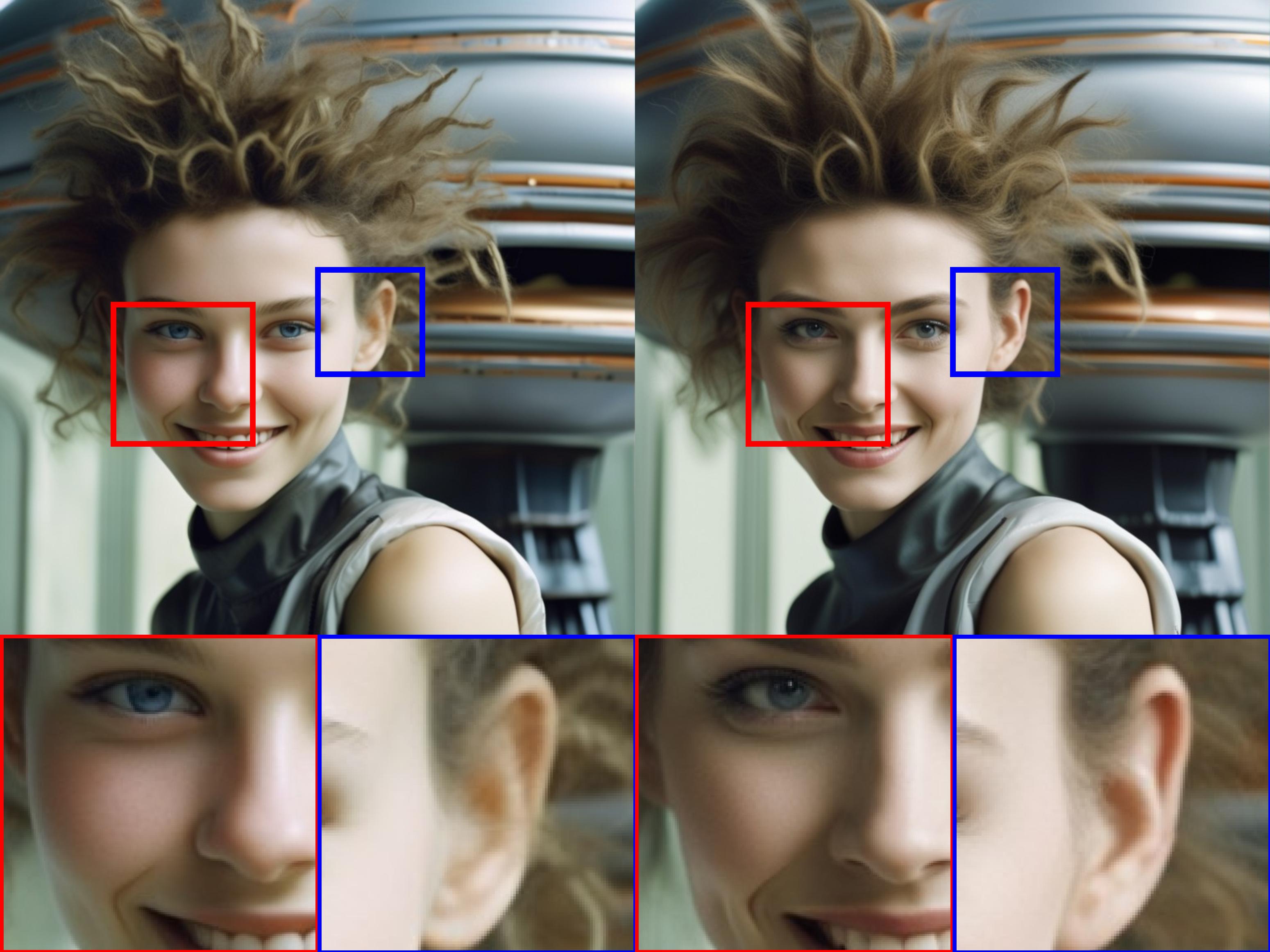}\\
     \vspace{5mm}
     \includegraphics[width=\textwidth]{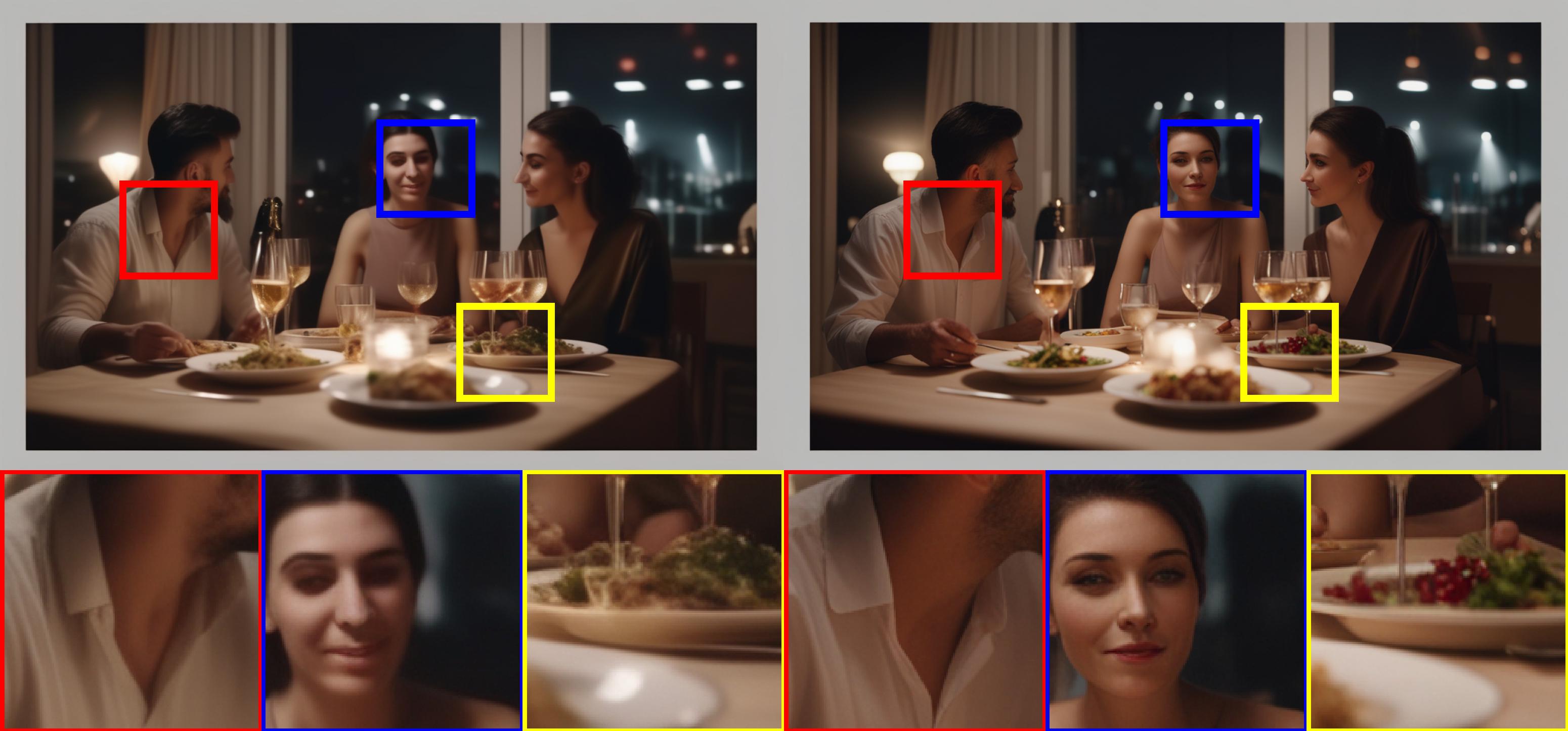}
    \caption{\modelname samples (with zoom-ins) without (left) and with (right) the refinement model discussed. Prompt: (\emph{top}) ``close up headshot, futuristic young woman, wild hair sly smile in front of gigantic UFO, dslr, sharp focus, dynamic composition'' (\emph{bottom}) ``Three people having dinner at a table at new years eve, cinematic shot, 8k''. Zoom-in for details.}
    \label{refinevsnorefineapp}
\end{figure}
}

\newcommand{\refinevsnorefinevthree}{
\begin{figure}[htbp] %
     \centering
     \includegraphics[width=\textwidth]{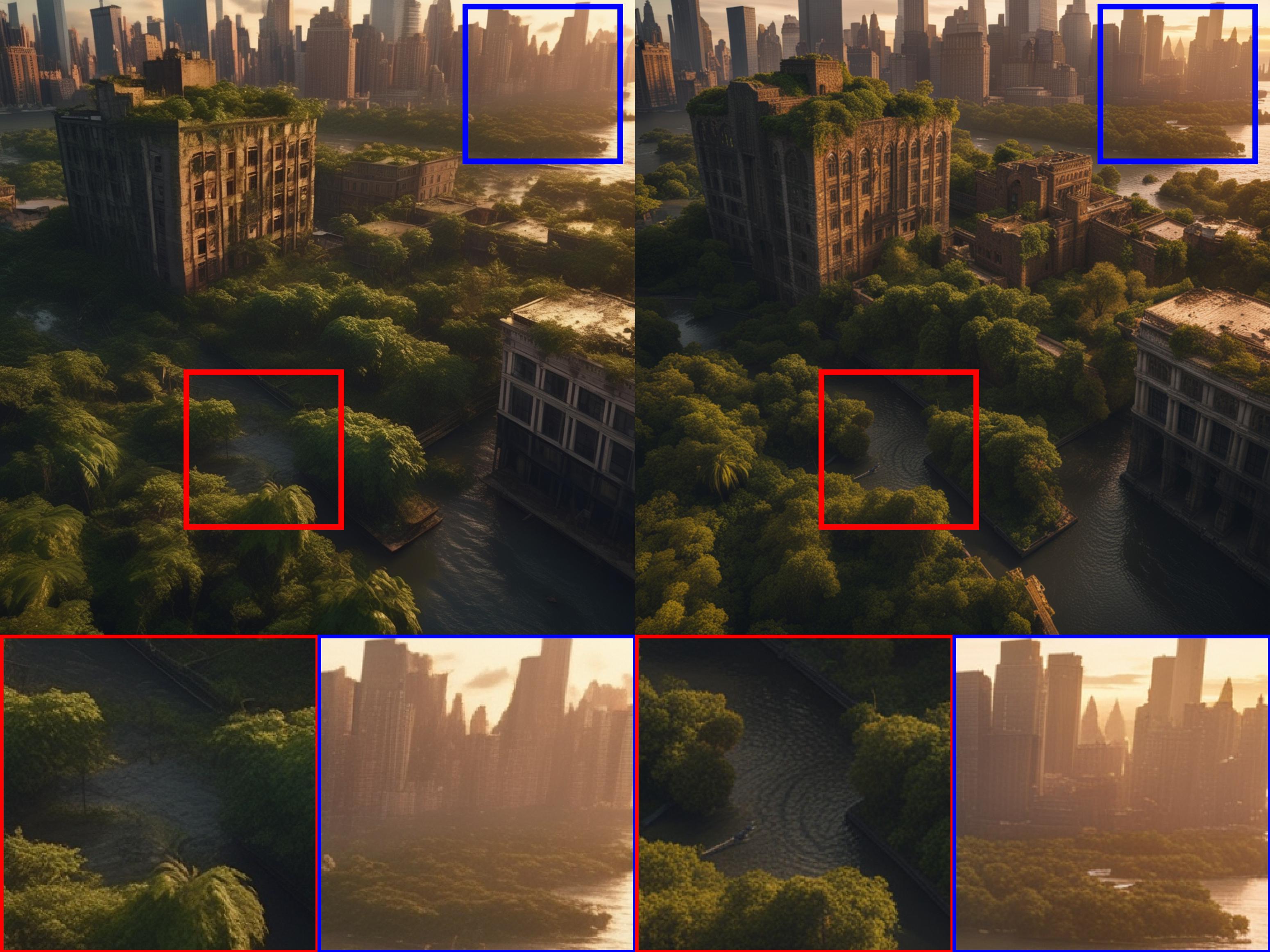}
    \caption{$1024^2$ samples (with zoom-ins) from \modelname without (left) and with (right) the refinement model discussed. Prompt: ``Epic long distance cityscape photo of New York City flooded by the ocean and overgrown buildings and jungle ruins in rainforest, at sunset, cinematic shot, highly detailed, 8k, golden light''. See~\Cref{refinevsnorefineapp} for additional samples.\vspace{-1em}}
    \label{fig:refinevsnorefine}
\end{figure}
}

\newcommand{\condcatcode}{
\centering
\begin{figure}[htbp]
\centering
\resizebox{.81\textwidth}{!}{%
\lstinputlisting[language=Python]{concat_conds_smaller.py}
}
\caption{\label{fig:cond_cat_code} Python code for concatenating the additional conditionings introduced in \Crefrange{subsec:archi_and_scale}{subsec:mar} along the channel dimension. }
\end{figure}
}

\newcommand{\sdoldcompapp}{
\renewcommand{\imwidth}{.47\textwidth}
\renewcommand{\impatha}[1]{img/comp_old_model/sd1-5/##1}
\renewcommand{\impathb}[1]{img/comp_old_model/sd2-1/##1}
\renewcommand{\impathc}[1]{img/comp_old_model/sdxl/##1}
\begin{figure}[htbp]
\centering
\begin{tabular}{c@{\hspace{5pt}}c@{\hspace{3pt}}c}
\toprule
& \footnotesize{\shortstack{\emph{'Vibrant portrait painting of Salvador Dalí} \\ \emph{with a robotic half face.'}}}  & \footnotesize{\shortstack{\emph{'A capybara made of voxels} \\ \emph{sitting in a field.'}}} \\
\midrule
\rotatebox[origin=l,y=1.2em]{90}{\large{\textbf{\sdshort 1-5}}} &
\includegraphics[width=\imwidth]{\impatha{dali_row}} &
\includegraphics[width=\imwidth]{\impatha{capybara_row}} \\ %

\rotatebox[origin=l,y=1.2em]{90}{\large{\textbf{\sdshort 2-1}}} &
\includegraphics[width=\imwidth]{\impathb{dali_row}} &
\includegraphics[width=\imwidth]{\impathb{capybara_row}} \\ %

\rotatebox[origin=l,y=1.2em]{90}{\large{\textbf{\modelname}}} &
\includegraphics[width=\imwidth]{\impathc{dali_row}} &
\includegraphics[width=\imwidth]{\impathc{capybara_row}} \\ %

\bottomrule
\toprule

& \scriptsize{\shortstack{\emph{'Cute adorable little goat, unreal engine,} \\ \emph{cozy interior lighting, art station, detailed’} \\ \emph{ digital painting, cinematic, octane rendering.'}}}  & \scriptsize{\shortstack{\emph{'A portrait photo of a kangaroo wearing an orange hoodie} \\ \emph{and blue sunglasses standing on the grass in front of the Sydney} \\ \emph{Opera House holding a sign on the chest that says "SDXL"!.'}}} \\
\midrule
\rotatebox[origin=l,y=1.2em]{90}{\large{\textbf{\sdshort 1-5}}} &
\includegraphics[width=\imwidth]{\impatha{goat_row}} &
\includegraphics[width=\imwidth]{\impatha{cangaroo_row}} \\ %

\rotatebox[origin=l,y=1.2em]{90}{\large{\textbf{\sdshort 2-1}}} &
\includegraphics[width=\imwidth]{\impathb{goat_row}} &
\includegraphics[width=\imwidth]{\impathb{cangaroo_row}} \\ %

\rotatebox[origin=l,y=1.2em]{90}{\large{\textbf{\modelname}}} &
\includegraphics[width=\imwidth]{\impathc{goat_row}} &
\includegraphics[width=\imwidth]{\impathc{cangaroo_row}} \\ %

\bottomrule
\end{tabular}
\caption{\label{fig:comp_old_model_app} Additional results for the comparison of the output of \modelname with previous versions of \sd. For each prompt, we show 3 random samples of the respective model for 50 steps of the DDIM sampler~\cite{song2020denoising} and cfg-scale $8.0$~\cite{ho2022classifier} 
}
\end{figure}
}

\newcommand{\sdoldcompapptwo}{
\renewcommand{\imwidth}{.47\textwidth}
\renewcommand{\impatha}[1]{img/comp_old_model/sd1-5/##1}
\renewcommand{\impathb}[1]{img/comp_old_model/sd2-1/##1}
\renewcommand{\impathc}[1]{img/comp_old_model/sdxl/##1}
\begin{figure}[htbp]

\centering
\begin{tabular}{c@{\hspace{5pt}}c@{\hspace{3pt}}c}
\toprule
& \footnotesize{\shortstack{\emph{'Monster Baba yaga house with in a forest,} \\ \emph{dark horror style, black and white.'}}}  & \footnotesize{\shortstack{\emph{'A young badger delicately sniffing a } \\ \emph{yellow rose, richly textured oil painting.'}}} \\
\midrule
\rotatebox[origin=l,y=1.2em]{90}{\large{\textbf{\sdshort 1-5}}} &
\includegraphics[width=\imwidth]{\impatha{baba_row}} &
\includegraphics[width=\imwidth]{\impatha{badger_row}} \\ %

\rotatebox[origin=l,y=1.2em]{90}{\large{\textbf{\sdshort 2-1}}} &
\includegraphics[width=\imwidth]{\impathb{baba_row}} &
\includegraphics[width=\imwidth]{\impathb{badger_row}} \\ %

\rotatebox[origin=l,y=1.2em]{90}{\large{\textbf{\modelname}}} &
\includegraphics[width=\imwidth]{\impathc{baba_row}} &
\includegraphics[width=\imwidth]{\impathc{badger_row}} \\ %

\bottomrule
\end{tabular}
\caption{\label{fig:comp_old_model_app2} Additional results for the comparison of the output of \modelname with previous versions of \sd. For each prompt, we show 3 random samples of the respective model for 50 steps of the DDIM sampler~\cite{song2020denoising} and cfg-scale $8.0$~\cite{ho2022classifier}.
}
\end{figure}
}

\definecolor{modelcolor}{rgb}{0.345, 0.471, 0.739}
\definecolor{compcolor}{rgb}{0.894, 0.576, 0.267}

\newcommand{\mjcompzero}{
\begin{figure}[htbp]
  \centering
  \includegraphics[trim=60 0 0 0,clip,width=\textwidth]{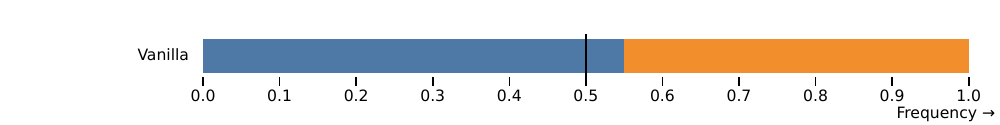}    
    \caption{Results from 17,153 user preference comparisons between \textcolor{modelcolor}{\modelname} v0.9 and \textcolor{compcolor}{Midjourney v5.1}, which was the latest version available at the time. The comparisons span all ``categories'' and ``challenges'' in the PartiPrompts (P2) benchmark. Notably, \textcolor{modelcolor}{\modelname} was favored 54.9\% of the time over \textcolor{compcolor}{Midjourney V5.1}. Preliminary testing indicates that the recently-released \textcolor{compcolor}{Midjourney V5.2} has lower prompt comprehension than its predecessor, but the laborious process of generating multiple prompts hampers the speed of conducting broader tests.}
  \label{fig:mjcomp_total}
\end{figure}
}

\newcommand{\mjcompuno}{
\begin{figure}[htbp]
  \centering
  \includegraphics[trim=20 0 0 0,clip,width=\textwidth]{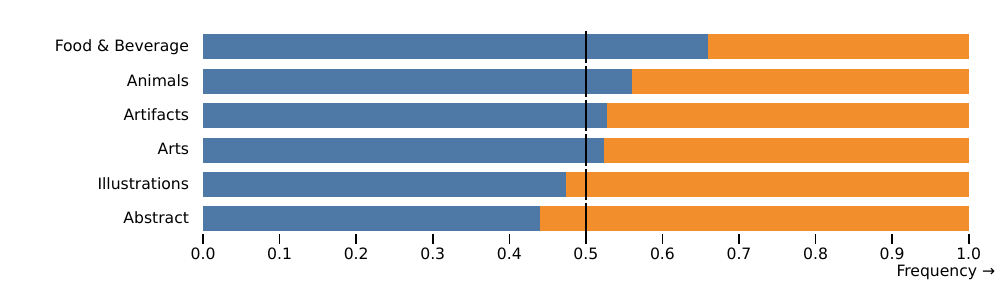}
  \caption{User preference comparison of \textcolor{modelcolor}{\modelname} (without refinement model) and \textcolor{compcolor}{Midjourney V5.1} across particular text categories. \textcolor{modelcolor}{\modelname} outperforms \textcolor{compcolor}{Midjourney V5.1} in all but two categories.}
  \label{fig:mjcomp_categories}
\end{figure}
}

\newcommand{\mjcompdue}{
\begin{figure}[htbp]
  \centering
  \includegraphics[width=\textwidth]{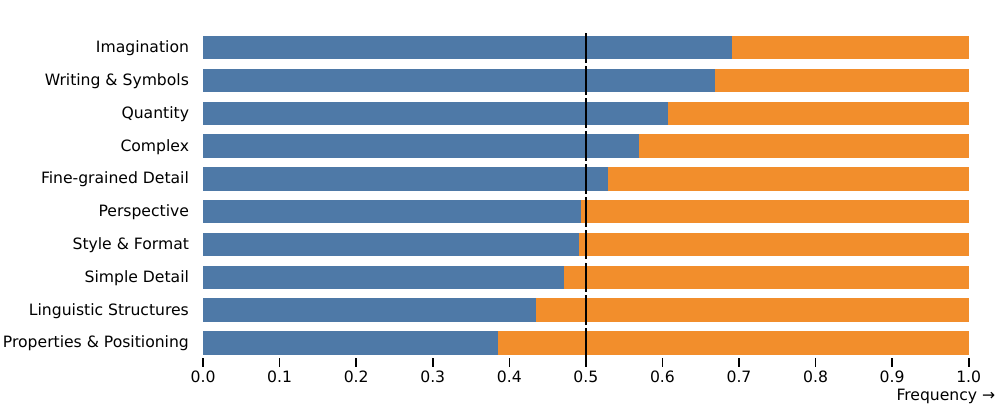}
  \caption{Preference comparisons of \textcolor{modelcolor}{\modelname} (with refinement model) to \textcolor{compcolor}{Midjourney V5.1} on complex prompts. \textcolor{modelcolor}{\modelname} either outperforms or is statistically equal to \textcolor{compcolor}{Midjourney V5.1} in 7 out of 10 categories.}
  \label{fig:mjcomp_challenges}
\end{figure}
}

\newcommand{\sdoldcomp}{
\renewcommand{\imwidth}{.47\textwidth}
\renewcommand{\impatha}[1]{img/comp_old_model/sd1-5/##1}
\renewcommand{\impathb}[1]{img/comp_old_model/sd2-1/##1}
\renewcommand{\impathc}[1]{img/comp_old_model/sdxl/##1}
\begin{figure}[htbp]
\vspace{-1em}
\centering
\begin{tabular}{c@{\hspace{5pt}}c@{\hspace{3pt}}c}
\toprule
& \footnotesize{\shortstack{\emph{'A propaganda poster depicting a cat dressed as french} \\ \emph{emperor
napoleon holding a piece of cheese.'}}} & \footnotesize{\shortstack{\emph{'a close-up of a fire spitting dragon,} \\ \emph{cinematic shot.'}}} \\ 
\midrule
\rotatebox[origin=l,y=1.2em]{90}{\large{\textbf{\sdshort 1-5}}} &
\includegraphics[width=\imwidth]{\impatha{catpoleon_row}} &
\includegraphics[width=\imwidth]{\impatha{dragon_row}} \\ %

\rotatebox[origin=l,y=1.2em]{90}{\large{\textbf{\sdshort 2-1}}} &
\includegraphics[width=\imwidth]{\impathb{catpoleon_row}} &
\includegraphics[width=\imwidth]{\impathb{dragon_row}} \\ %

\rotatebox[origin=l,y=1.2em]{90}{\large{\textbf{\modelname}}} &
\includegraphics[width=\imwidth]{\impathc{catpoleon_row}} &
\includegraphics[width=\imwidth]{\impathc{dragon_row}} \\ %
\bottomrule
\end{tabular}
\caption{\label{fig:comp_old_model} Comparison of the output of \modelname with previous versions of \sd. For each prompt, we show 3 random samples of the respective model for 50 steps of the DDIM sampler~\cite{song2020denoising} and cfg-scale $8.0$~\cite{ho2022classifier}. Additional samples in \Cref{fig:comp_old_model_app}. \vspace{-1.0em}
}
\end{figure}
}

\newcommand{\failureplot}{
\renewcommand{\imwidth}{.48\textwidth}
\renewcommand{\impath}[1]{img/failure/##1}
\begin{figure}[htbp]

\centering
\begin{tabular}{c@{\hspace{5pt}}c}
\toprule
\footnotesize{\shortstack{\emph{'A close up of a handpalm} \\ \emph{with leaves growing from it.'}}}   &
\footnotesize{\shortstack{\emph{'An empty fireplace with a television above it.} \\ 
\emph{The TV shows a lion hugging a giraffe.'}}}   \\

\midrule

\includegraphics[width=\imwidth]{\impath{leave_handpalm}} &
\includegraphics[width=\imwidth]{\impath{lion_giraffe}} \\ %

\midrule
\footnotesize{\emph{'A grand piano with a white bench.'}} &
\footnotesize{\shortstack{\emph{'Three quarters view of a rusty old red pickup} \\ \emph{truck with white doors and a smashed windshield.'}}}   \\

\midrule

\includegraphics[width=\imwidth]{\impath{piano}} &
\includegraphics[width=\imwidth]{\impath{truck}} \\ %

\bottomrule
\end{tabular}
\caption{\label{fig:failure_cases} Failure cases of \modelname despite large improvements compared to previous versions of \sd, the model sometimes still struggles with very complex prompts involving detailed spatial arrangements and detailed descriptions (e.g. top left example). Moreover, hands are not yet always correctly generated (e.g. top left) and the model sometimes suffers from two concepts bleeding into one another (e.g. bottom right example).  All examples are random samples generated with 50 steps of the DDIM sampler~\cite{song2020denoising} and cfg-scale $8.0$~\cite{ho2022classifier}.
}
\end{figure}
}

\newcommand{\modelarchcomp}{
\begin{table}
\caption{\label{tab:modelarchcomp} Comparison of \modelname and older \sd models.}
\begin{tabular}{lccc}
\toprule
Model & \modelname & SD 1.4/1.5 & SD 2.0/2.1 \\
\midrule
\# of UNet params & 2.6B & 860M & 865M \\
Transformer blocks & [0, 2, 10] & [1, 1, 1, 1] & [1, 1, 1, 1]\\
Channel mult. & [1, 2, 4] & [1, 2, 4, 4] & [1, 2, 4, 4] \\
Text encoder & CLIP ViT-L \& OpenCLIP ViT-bigG  & CLIP ViT-L & OpenCLIP ViT-H  \\
Context dim. & 2048 & 768 & 1024 \\
Pooled text emb. & OpenCLIP ViT-bigG & N/A & N/A \\
\bottomrule
\end{tabular}
\end{table}
}

\newcommand{\aecomp}{
\begin{wraptable}{r}{.5\textwidth}
\vspace{-3em}
\centering
\caption{\label{tab:aecomp} Autoencoder reconstruction performance on the COCO2017~\cite{lin2015microsoft} validation split, images of size $256\times 256$ pixels. Note: \sd 2.x uses an improved version of \sd 1.x's autoencoder, where the decoder was finetuned with a reduced weight on the perceptual loss~\cite{zhang2018unreasonable}, and used more compute. Note that our new autoencoder is trained from scratch.}
\resizebox{.5\textwidth}{!}{%
\begin{tabular}{lcccc}
\toprule
model & PNSR $\uparrow$ & SSIM $\uparrow$ & LPIPS $\downarrow$ & rFID $\downarrow$ \\
\midrule
\modelname-VAE & $\fat{24.7}$ & $\fat{0.73}$ & $\fat{0.88}$ & $\fat{4.4}$ \\
\sdshort-VAE 1.x & 23.4 & 0.69 & 0.96 & 5.0 \\ 
\sdshort-VAE 2.x & 24.5 & 0.71 & 0.92 & 4.7 \\ 
\bottomrule
\end{tabular}}
\vspace{-1em}
\end{wraptable}
}

\newcommand{\sizecondtable}{
\begin{wraptable}{r}{.4\textwidth}
\vspace{-1.2em}
\centering
\caption{\label{tab:size-cond} Conditioning on the original spatial size of the training examples improves performance on class-conditional ImageNet~\cite{deng2009imagenet} on $512^2$ resolution. 
}
\begin{tabular}{lcc}
\toprule
model &  FID-5k $\downarrow$ & IS-5k $\uparrow$ \\
\midrule 
\emph{CIN-512-only}  & 43.84 & 110.64 \\ 
\emph{CIN-nocond} & 39.76 &  211.50 \\
\emph{CIN-size-cond} & \textbf{36.53} & \textbf{215.34} \\  
\bottomrule
\end{tabular}
\vspace{-.5em}
\end{wraptable}
}
\newcommand{\condaug}{
\begin{algorithm}
\caption{\label{alg:cond} Conditioning pipeline for size- and crop-conditioning}
\begin{algorithmic}
\Require Training dataset of images $\boldsymbol{\mathcal{D}}$, target image size for training $\boldsymbol{s} = (h_{\mathrm{tgt}}, w_{\mathrm{tgt}})$ 
\Require Resizing function $\boldsymbol{R}$, cropping function function $\boldsymbol{C}$
\Require Model train step $\boldsymbol{T}$
\State converged $\gets$ False
\While{not converged}
\State $x \sim \boldsymbol{\mathcal{D}}$
\State $w_{\text{\tiny{original}}} \gets \mathrm{width}(x)$
\vspace{2pt}
\State $h_{\text{\tiny{original}}} \gets \mathrm{height}(x)$
\vspace{2pt}
\State $\csize \gets (h_{\text{\tiny{original}}},w_{\text{\tiny{original}}})$
\vspace{2pt}
\State $x \gets \boldsymbol{R}(x, \boldsymbol{s})$ \Comment{resize smaller image size to target size $\boldsymbol{s}$}
\vspace{2pt}
\If{$h_{\text{\tiny{original}}} \leq w_{\text{\tiny{original}}}$}
    \vspace{2pt}
    \State $c_{\text{\tiny{left}}} \sim \boldsymbol{\mathcal{U}}(0,\mathrm{width}(x) - s_w)$ \Comment{sample $c_{\text{\tiny{left}}}$ from discrete uniform distribution}
    \vspace{2pt}
    \State $c_{\text{\tiny{top}}} = 0$
    \vspace{2pt}
\ElsIf{$h_{\text{\tiny{original}}} > w_{\text{\tiny{original}}}$}
    \vspace{2pt}
    \State $c_{\text{\tiny{top}}} \sim \boldsymbol{\mathcal{U}}(0,\mathrm{height}(x) - s_h)$ \Comment{sample $c_{\text{\tiny{top}}}$ from discrete uniform distribution}
    \vspace{2pt}
    \State $c_{\text{\tiny{left}}} = 0$
    \vspace{2pt}
\EndIf
\State $\ccrop \gets \left(c_{\text{\tiny{top}}}, c_{\text{\tiny{left}}}\right)$
\vspace{2pt}
\State $x \gets \boldsymbol{C}(x, \boldsymbol{s}, \ccrop)$ \Comment{crop image to size $\boldsymbol{s}$ with top-left coordinate 
$\left(c_{\text{\tiny{top}}}, c_{\text{\tiny{left}}}\right)$}
\vspace{2pt}
\State converged $\gets \boldsymbol{T}(x, \csize, \ccrop)$ \Comment{train model conditioned on $\csize$ and $\ccrop$}
\EndWhile
\end{algorithmic}
\end{algorithm}
}

\begin{document}
\vspace{-2em}
\enlargethispage\baselineskip
\vspace{-15em}
\maketitle
\vspace{-2em}
\begin{tabular}{c}
    \small Stability AI, Applied Research \\
    \addlinespace[5pt]
    \scriptsize {\emph{Code}: \url{https://github.com/Stability-AI/generative-models} \quad \emph{Model weights}: \url{https://huggingface.co/stabilityai/}}
\end{tabular}
\teaser
\vspace{-1em}
\begin{abstract}
We present \modelname, a latent diffusion model for text-to-image synthesis. 
Compared to previous versions of \sd, \modelname leverages a three times larger UNet backbone: The increase of model parameters is mainly due to more attention blocks and a larger cross-attention context as \modelname uses a second text encoder. 
We design multiple novel conditioning schemes and train \modelname on multiple aspect ratios. 
We also introduce a \emph{refinement model} which is 
used to improve the visual fidelity of samples generated by \modelname using a post-hoc \emph{image-to-image} technique.
We demonstrate that \modelname shows drastically improved performance compared to previous versions of \sd and achieves results competitive with those of black-box state-of-the-art image generators.
In the spirit of promoting open research and fostering transparency in large model training and evaluation, we provide access to code and model weights.

\end{abstract}

\section{Introduction}
\label{sec:intro}
The last year has brought enormous leaps in deep generative modeling across various data domains, such as natural language~\cite{touvron2023llama}, audio~\cite{huang2023make}, and visual media~\citep{rombach2021high,ramesh2022hierarchical,saharia2022photorealistic,singer2022make,ho2022imagen,blattmann2023align, esser2023structure}.
In this report, we focus on the latter and unveil \modelname, a drastically improved version of \sd. 
\sd is a latent text-to-image diffusion model (DM) which serves as the foundation for an array of recent advancements in, e.g., 
3D classification~\cite{shen2023diffclip}, controllable image editing~\cite{zhang2023adding}, 
image personalization~\cite{gal2022image}, synthetic data augmentation~\cite{stockl2022evaluating}, 
graphical user interface prototyping~\cite{wei2023boosting}, etc. Remarkably, the scope of applications has been extraordinarily extensive, encompassing fields as diverse as music generation~\cite{Forsgren_Martiros_2022} and reconstructing images from fMRI brain scans~\cite{takagi2023high}.

User studies demonstrate that \modelname consistently surpasses all previous versions of \sd by a significant margin (see~\Cref{fig:userstudyandmodel}).
In this report, we present 
the design choices which lead to this boost in performance encompassing \emph{i)} a 3$\times$ larger UNet-backbone 
compared to previous \sd models (\Cref{subsec:archi_and_scale}), \emph{ii)} 
two simple yet effective additional conditioning techniques (\Cref{subsec:condtricks}) which do not require any form of additional supervision, and \emph{iii)} a separate diffusion-based refinement model which applies a noising-denoising process~\cite{meng2021sdedit} to the latents produced by \modelname to improve the visual quality of its samples (\Cref{subsec:putting}).  

A major concern in the field of visual media creation is that while black-box-models 
are often 
recognized as state-of-the-art, the opacity of their architecture 
prevents faithfully assessing and validating their performance. This lack of transparency hampers reproducibility, stifles innovation, and prevents the community from building upon these models to further the progress of science and art. Moreover, 
these closed-source strategies make it challenging to assess the biases and limitations of these models in an impartial and objective way, which is crucial for their responsible and ethical deployment. With \modelname we are releasing an \emph{open} model that achieves competitive performance with black-box image generation models (see ~\Cref{fig:mjcomp_categories}~\&~\Cref{fig:mjcomp_challenges}).

\section{Improving \sd}
In this section we present our improvements for the \sd architecture. These are modular, and can be used individually or together to extend any model. Although the following strategies are implemented as extensions to latent diffusion models (LDMs)~\cite{rombach2021high}, most of them are also applicable to their pixel-space counterparts.
\userstudyandmodel
\subsection{Architecture \& Scale}
\label{subsec:archi_and_scale}
\modelarchcomp

Starting with the seminal works \citet{ho2020denoising} and \citet{song2020score}, which demonstrated that DMs are powerful generative models for image synthesis, the convolutional UNet~\cite{ronneberger2015u} architecture has been the dominant architecture for diffusion-based image synthesis. However, with the development of foundational DMs~\cite{saharia2022photorealistic,ramesh2022hierarchical,rombach2021high}, the underlying architecture has constantly evolved: from adding self-attention and improved upscaling layers~\cite{dhariwal2021diffusion}, over cross-attention for text-to-image synthesis~\cite{rombach2021high}, to pure transformer-based architectures~\cite{peebles2022scalable}.

We follow this trend and, following~\citet{hoogeboom2023simple}, shift the bulk of the transformer computation to lower-level features in the UNet. In particular, and in contrast to the original \sd architecture, we use a heterogeneous distribution of transformer blocks within the UNet: For efficiency reasons, we omit the transformer block at the highest feature level, use 2 and 10 blocks at the lower levels, and remove the lowest level ($8 \times$ downsampling) in the UNet altogether --- see~\Cref{tab:modelarchcomp} for a comparison between the architectures of \sd 1.x \& 2.x and \modelname. We opt for a more powerful pre-trained text encoder that we use for text conditioning. Specifically, we use OpenCLIP ViT-bigG~\cite{ilharco_gabriel_2021_5143773} in combination with CLIP ViT-L~\cite{radford2021learning}, where we concatenate the penultimate text encoder outputs along the channel-axis~\cite{balaji2022ediffi}. Besides using cross-attention layers to condition the model on the text-input, we follow ~\cite{nichol2021glide} and additionally condition the model on the pooled text embedding from the OpenCLIP model. These changes result in a model size of 2.6B parameters in the UNet, see Tab.~\ref{tab:modelarchcomp}. The text encoders have a total size of 817M parameters.

\subsection{Micro-Conditioning}
\label{subsec:condtricks}
\sizedist
\paragraph{Conditioning the Model on Image Size}
A notorious shortcoming of the LDM paradigm~\cite{rombach2021high} is the fact that training a model requires a \emph{minimal image size}, due to its two-stage architecture. The two main approaches to tackle this problem are either to discard all training images below a certain minimal resolution (for example, \sd 1.4/1.5 discarded all images with any size below 512 pixels), or, alternatively, upscale images that are too small. However, depending on the desired image resolution, the former method can lead to significant portions of the training data being discarded, what will likely lead to a loss in performance and hurt generalization. We visualize such effects in~\Cref{fig:size_dist} for the dataset on which \modelname was pretrained. For this particular choice of data, discarding all samples below our pretraining resolution of $256^2$ pixels would lead to a significant 39\% of discarded data. The second method, on the other hand, usually introduces upscaling artifacts which may leak into the final model outputs, causing, for example, blurry samples.

Instead, we propose to condition the UNet model on the original image resolution, which is trivially available during training.
In particular, we provide the original (i.e., before any rescaling) height and width of the images as an additional conditioning to the model $\csize = (h_{\text{original}}, w_{\text{original}})$. Each component is independently embedded using a Fourier feature encoding, and these encodings are concatenated into a single vector that we feed into the model by adding it to the timestep embedding~\cite{dhariwal2021diffusion}. 

\sizecondvtwo

At inference time, a user can then set the desired \emph{apparent resolution} of the image via this \emph{size-conditioning}. Evidently (see \Cref{fig:sizecond}), the model has learned to associate the conditioning $c_{\text{size}}$ with resolution-dependent image features, which can be leveraged to modify the appearance of an output corresponding to a given prompt. Note that for the visualization shown in \Cref{fig:sizecond}, we visualize samples generated by the $512 \times 512$ model (see \Cref{subsec:putting} for details), since the effects of the size conditioning are less clearly visible after the subsequent multi-aspect (ratio) finetuning which we use for our final \modelname model.  

\sizecondtable
We quantitatively assess the effects of this simple but effective conditioning technique by training and evaluating three LDMs on class conditional ImageNet~\cite{deng2009imagenet} at spatial size $512^2$: For the first model (\emph{CIN-512-only}) we discard all training examples with at least one edge smaller than $512$ pixels what results in a train dataset of only 70k images. For \emph{CIN-nocond} we use all training examples but without size conditioning. This additional conditioning is only used for \emph{CIN-size-cond}. After training we generate 5k samples with 50 DDIM steps~\citep{song2020denoising} and (classifier-free) guidance scale of 5~\citep{ho2022classifier} for every model and compute IS~\cite{salimans2016improved} and FID~\cite{heusel2017gans} (against the full validation set). For \emph{CIN-size-cond} we generate samples always conditioned on $\csize=(512,512)$. \Cref{tab:size-cond} summarizes the results and verifies that \emph{CIN-size-cond} improves upon the baseline models in both metrics. We attribute the degraded performance of \emph{CIN-512-only} to bad generalization due to overfitting on the small training dataset while the effects of a mode of blurry samples in the sample distribution of \emph{CIN-nocond} result in a reduced FID score. Note that, although we find these classical quantitative scores not to be suitable for evaluating the performance of foundational (text-to-image) DMs~\cite{saharia2022photorealistic,ramesh2022hierarchical,rombach2021high} (see \Cref{suppsec:fid_app}), they remain reasonable metrics on ImageNet as the neural backbones of FID and IS have been trained on ImageNet itself. 

\paragraph{Conditioning the Model on Cropping Parameters}
\sdoldcomp
The first two rows of \Cref{fig:comp_old_model} illustrate a typical failure mode of previous \sdshort models: Synthesized objects can be cropped, such as the cut-off head of the cat in the left examples for \sdshort 1-5 and \sdshort 2-1.
An intuitive explanation for this behavior is the use of \emph{random cropping} during training of the model: As collating a batch in DL frameworks such as PyTorch~\cite{paszke2019pytorch} requires tensors of the same size, a typical processing pipeline is to (i) resize an image such that the shortest size matches the desired target size, followed by (ii) randomly cropping the image along the longer axis. 
While random cropping is a natural form of data augmentation, it can leak into the generated samples, causing the malicious effects shown above.

To fix this problem, we propose another simple yet effective conditioning method: During dataloading, we uniformly sample crop coordinates $c_{\text{top}}$ and $c_{\text{left}}$ (integers specifying the amount of pixels cropped from the top-left corner along the height and width axes, respectively) and feed them into the model as conditioning parameters via Fourier feature embeddings, similar to the size conditioning described above. The concatenated embedding $\ccrop$ is then used as an additional conditioning parameter. We emphasize that this technique is not limited to LDMs and could be used for any DM.
Note that crop- and size-conditioning can be readily combined. In such a case, we concatenate the feature embedding along the channel dimension, before adding it to the timestep embedding in the UNet. \Cref{alg:cond} illustrates how we sample $\ccrop$ and $\csize$ during training if such a combination is applied.
\condaug

Given that in our experience large scale datasets are, on average, object-centric, we %
set $\left(c_{\text{top}}, c_{\text{left}}\right) = \left(0, 0\right)$ during inference and thereby obtain object-centered samples from the trained model.

\cropcondvtwo
See Fig.~\ref{fig:cropcond} for an illustration: By tuning $\left(c_{\text{top}}, c_{\text{left}}\right)$, we can successfully \emph{simulate} the amount of cropping during inference. This is a form of \emph{conditioning-augmentation}, and has been used in various forms with autoregressive~\cite{jun2020distribution} models, %
and more recently with diffusion models~\cite{karras2022elucidating}. 

While other methods like data bucketing~\cite{novelai2023novelai} successfully tackle the same task, we still benefit from cropping-induced data augmentation, while making sure that it does not leak into the generation process - we actually use it to our advantage to gain more control over the image synthesis process. Furthermore, it is easy to implement and can be applied in an online fashion during training, without additional data preprocessing.

\subsection{Multi-Aspect Training}
\label{subsec:mar}
Real-world datasets include images of widely varying sizes and aspect-ratios (c.f. \cref{fig:size_dist}) While the common output resolutions for text-to-image models are square images of $512\times512$ or $1024\times1024$ pixels, we argue that this is a rather unnatural choice, given the widespread distribution and use of landscape (e.g., 16:9) or portrait format screens. %

Motivated by this, we finetune our model to handle multiple aspect-ratios simultaneously: We follow common practice \cite{novelai2023novelai} and partition the data into buckets of different aspect ratios, where we keep the pixel count as close to $1024^2$ pixels as possibly, varying height and width accordingly in multiples of 64. A full list of all aspect ratios used for training is provided in \Cref{supsubsec:marlist}. During optimization, a training batch is composed of images from the same bucket, and we alternate between bucket sizes for each training step. Additionally, the model receives the bucket size (or, \emph{target size}) as a conditioning, represented as a tuple of integers $\car=(h_{\text{tgt}}, w_{\text{tgt}})$ which are embedded into a Fourier space in analogy to the size- and crop-conditionings described above.%

In practice, we apply multi-aspect training as a finetuning stage after pretraining the model at a fixed aspect-ratio and resolution and combine it with the conditioning techniques introduced in \Cref{subsec:condtricks} via concatenation along the channel axis. \Cref{fig:cond_cat_code} in \Cref{supsec:cond_pseudo_code} provides \verb|python|-code for this operation. Note that crop-conditioning and multi-aspect training are complementary operations, and crop-conditioning then only works within the bucket boundaries (usually 64 pixels). For ease of implementation, however, we opt to keep this control parameter for multi-aspect models.

\subsection{Improved Autoencoder}
\label{subsec:ae}
\aecomp
\sd is a \emph{LDM}, operating in a pretrained, learned (and fixed) latent space of an autoencoder. While the bulk of the semantic composition is done by the LDM~\cite{rombach2021high}, we can improve \emph{local}, high-frequency details in generated images by improving the autoencoder. To this end, we train the same autoencoder architecture used for the original \sd at a larger batch-size (256 vs 9) and additionally track the weights with an exponential moving average. The resulting autoencoder outperforms the original model in all evaluated reconstruction metrics, see~\Cref{tab:aecomp}. We use this autoencoder for all of our experiments.

\subsection{Putting Everything Together}
\label{subsec:putting}
We train the final model, \modelname, in a multi-stage procedure. \modelname uses the autoencoder from~\Cref{subsec:ae} and a discrete-time diffusion schedule~\cite{ho2020denoising,sohl2015deep} with \num{1000} steps. 
First, we pretrain a base model (see~\Cref{tab:modelarchcomp}) on an internal dataset whose height- and width-distribution is visualized in Fig.~\ref{fig:size_dist} for \num{600000} optimization steps at a resolution of $256\times 256$ pixels and a batch-size of \num{2048}, using size- and crop-conditioning as described in Sec.~\ref{subsec:condtricks}. 
We continue training on $512 \times 512$ pixel images for another \num{200000} optimization steps, and finally utilize multi-aspect training (\Cref{subsec:mar}) in combination with an offset-noise~\cite{guttenberg2023diffusion, lin2023common} level of \num{0.05} to train the model on different aspect ratios (\Cref{subsec:mar}, \Cref{supsubsec:marlist}) of $\sim$ $1024 \times 1024$ pixel area.

\paragraph{Refinement Stage}
Empirically, we find that the resulting model sometimes yields samples of low local quality, see \Cref{fig:refinevsnorefine}. 
To improve sample quality, we train a separate LDM in the same latent space, which is specialized on high-quality, high resolution data and employ a noising-denoising process as introduced by \emph{SDEdit}~\cite{meng2021sdedit} on the samples from the base model. We follow \cite{balaji2022ediffi} and specialize this refinement model on the first 200 (discrete) noise scales.
During inference, we render latents from the base \modelname, and directly diffuse and denoise them in latent space with the refinement model (see \Cref{fig:userstudyandmodel}), using the same text input. We note that this step is optional, but improves sample quality for detailed backgrounds and human faces, as demonstrated in \Cref{fig:refinevsnorefine} and \Cref{refinevsnorefineapp}.

To assess the performance of our model (with and without refinement stage), we conduct a user study, and let users pick their favorite generation from the following four models: \modelname, \modelname (with refiner), \sd 1.5 and \sd 2.1. The results demonstrate the \modelname with the refinement stage is the highest rated choice, and outperforms \sd 1.5 \& 2.1 by a significant margin (win rates: \modelname w/ refinement: $48.44\%$, \modelname base: $36.93\%$, \sd 1.5: $7.91\%$, \sd 2.1: $6.71\%$). See~\Cref{fig:userstudyandmodel}, which also provides an overview of the full pipeline.
However, when using classical performance metrics such as FID and CLIP scores the improvements of \modelname over previous methods are not reflected as shown in \Cref{fig:fid+vs_clip} and discussed in \Cref{suppsec:fid_app}. This aligns with and further backs the findings of~\citet{kirstain2023pick}.

\refinevsnorefinevthree

\section{Future Work}
\enlargethispage{2\baselineskip}
This report presents a preliminary analysis of improvements to the foundation model \sd for text-to-image synthesis. While we achieve significant improvements in synthesized image quality, prompt adherence and composition, in the following, we discuss a few aspects for which we believe the model may be improved further: 

\begin{itemize}
\item Single stage: Currently, we generate the best samples from \modelname using a two-stage approach with an additional refinement model. This results in having to load two large models into memory, hampering accessibility and sampling speed. Future work should investigate ways to provide a single stage of equal or better quality.

\item Text synthesis: While the scale and the larger text encoder (OpenCLIP ViT-bigG~\cite{ilharco_gabriel_2021_5143773}) help to improve the text rendering capabilities over previous versions of \sd, incorporating byte-level tokenizers~\cite{xue2022byt5,liu2023characteraware} or simply scaling the model to larger sizes~\cite{yu2022scaling,saharia2022photorealistic} may further improve text synthesis.

\item Architecture: During the exploration stage of this work, we briefly experimented with transformer-based architectures such as UViT~\cite{hoogeboom2023simple} and DiT~\cite{peebles2022scalable}, but found no immediate benefit. We remain, however, optimistic that a careful hyperparameter study will eventually enable scaling to much larger transformer-dominated architectures.

\item Distillation: While our improvements over the original \sd model are significant, they come at the price of increased inference cost (both in VRAM and sampling speed). Future work will thus focus on decreasing the compute needed for inference, and increased sampling speed, for example through guidance-~\cite{meng2023distillation}, knowledge-~\cite{dockhorn2023distilling,kim2023architectural,li2023snapfusion} and progressive distillation~\cite{salimans2022progressive,berthelot2023tract,meng2023distillation}.

\item Our model is trained in the discrete-time formulation of \cite{ho2020denoising}, and requires \emph{offset-noise}~\cite{guttenberg2023diffusion, lin2023common} for aesthetically pleasing results. The EDM-framework of \citet{karras2022elucidating} is a promising candidate for future model training, as its formulation in continuous time allows for increased sampling flexibility and does not require noise-schedule corrections.
\end{itemize}

\clearpage
\newpage

\appendix
\begin{center}
\Huge\textbf{Appendix}
\end{center}

\section{Acknowledgements}
We thank all the folks at StabilityAI who worked on comparisons, code, etc, in particular: Alex Goodwin, Benjamin Aubin, Bill Cusick, Dennis Nitrosocke Niedworok, Dominik Lorenz, Harry Saini, Ian Johnson, Ju Huo, Katie May, Mohamad Diab, Peter Baylies, Rahim Entezari, Yam Levi, Yannik Marek, Yizhou Zheng. We also thank ChatGPT for providing writing assistance.

\section{Limitations}
\failureplot
While our model has demonstrated impressive capabilities in generating realistic images and synthesizing complex scenes, it is important to acknowledge its inherent limitations. Understanding these limitations is crucial for further improvements and ensuring responsible use of the technology.

Firstly, the model may encounter challenges when synthesizing intricate structures, such as human hands (see \Cref{fig:failure_cases}, top left). Although it has been trained on a diverse range of data, the complexity of human anatomy poses a difficulty in achieving accurate representations consistently. This limitation suggests the need for further scaling and training techniques specifically targeting the synthesis of fine-grained details. A reason for this occurring might be that hands and similar objects appear with very high variance in photographs and it is hard for the model to extract the knowledge of the real 3D shape and physical limitations in that case.

Secondly, while the model achieves a remarkable level of realism in its generated images, it is important to note that it does not attain perfect photorealism. Certain nuances, such as subtle lighting effects or minute texture variations, may still be absent or less faithfully represented in the generated images. This limitation implies that caution should be exercised when relying solely on model-generated visuals for applications that require a high degree of visual fidelity.

Furthermore, the model's training process heavily relies on large-scale datasets,
which can inadvertently introduce social and racial biases. As a result, the model may inadvertently exacerbate these biases when generating images or inferring visual attributes. %

In certain cases where samples contain multiple objects or subjects, the model may exhibit a phenomenon known as ``concept bleeding''.
This issue manifests as the unintended merging or overlap of distinct visual elements.
For instance, in \Cref{fig:comp_old_model_app}, an orange sunglass is observed, which indicates an instance of concept bleeding from the orange sweater.
Another case of this can be seen in \Cref{fig:comparetoif}, the penguin is supposed to have a ``blue hat'' and ``red gloves'', but is instead generated with blue gloves and a red hat.
Recognizing and addressing such occurrences is essential for refining the model's ability to accurately separate and represent individual objects within complex scenes.
The root cause of this may lie in the used pretrained text-encoders: firstly, they are trained to compress all information into a single token, so they may fail at binding only the right attributes and objects, \citet{feng2023trainingfree} mitigate this issue by explicitly encoding word relationships into the encoding. Secondly, the contrastive loss may also contribute to this, since negative examples with a different binding are needed within the same batch~\cite{rameshdalle2}.

Additionally, while our model represents a significant advancement over previous iterations of \sdshort, it still encounters difficulties when rendering long, legible text. Occasionally, the generated text may contain random characters or exhibit inconsistencies, as illustrated in \Cref{fig:comparetoif}.
Overcoming this limitation requires further investigation and development of techniques that enhance the model's text generation capabilities, particularly for extended textual content --- see for example the work of \citet{liu2023characteraware}, who propose to enhance text rendering capabilities via character-level text tokenizers. Alternatively, scaling the model does further improve text synthesis~\cite{yu2022scaling,saharia2022photorealistic}.

In conclusion, our model exhibits notable strengths in image synthesis, but it is not exempt from certain limitations. The challenges associated with synthesizing intricate structures, achieving perfect photorealism, further addressing biases, mitigating concept bleeding, and improving text rendering highlight avenues for future research and optimization.

\newpage 
\section{Diffusion Models}
In this section, we give a concise summary of DMs. We consider the continuous-time DM framework~\cite{song2020score} and follow the presentation of~\citet{karras2022elucidating}. 
Let $\pdata(\rvx_0)$ denote the data distribution and let $p(\rvx; \sigma)$ be the distribution obtained by adding i.i.d. $\sigma^2$-variance Gaussian noise to the data.
For sufficiently large $\sigma_{\mathrm{max}}$, $p(\rvx; \sigma_{\mathrm{max}^2})$ is almost indistinguishable from $\sigma^2_{\mathrm{max}}$-variance Gaussian noise. Capitalizing on this observation, DMs sample high variance Gaussian noise $\rvx_M \sim\gN\left(\bm{0}, \sigma_{\mathrm{max}^2}\right)$ and sequentially denoise $\rvx_M$ into $\rvx_i\sim p(\rvx_i; \sigma_i)$, $i \in \{0,\dots,M\}$, with $\sigma_{i} < \sigma_{i+1}$ and $\sigma_M=\sigma_{\mathrm{max}}$. For a well-trained DM and $\sigma_0=0$ the resulting $\rvx_0$ is distributed according to the data. 

\textbf{Sampling.} In practice, this iterative denoising process explained above can be implemented through the numerical simulation of the \emph{Probability Flow} ordinary differential equation (ODE)~\cite{song2020score}
\begin{align} \label{eq:probability_flow_ode}
    d\rvx = -\dot \sigma(t) \sigma(t) \nabla_\rvx \log p(\rvx; \sigma(t)) \, dt,
\end{align}
where $\nabla_\rvx \log p(\rvx; \sigma)$ is the \emph{score function}~\cite{hyvarinen2005estimation}. 
The schedule $\sigma(t) \colon [0, 1] \to \R_+$ is user-specified and  $\dot \sigma(t)$ denotes the time derivative of $\sigma(t)$. 
Alternatively, we may also numerically simulate a stochastic differential equation (SDE)~\citep{song2020score,karras2022elucidating}:
\begin{align} \label{eq:diffusion_sde}
    d\rvx = &\underbrace{- \dot \sigma(t) \sigma(t) \nabla_\rvx \log p(\rvx; \sigma(t)) \, dt}_{\text{Probability Flow ODE; see~\Cref{eq:probability_flow_ode}}} - \underbrace{\beta(t) \sigma^2(t) \nabla_\rvx \log p(\rvx; \sigma(t)) \, dt + \sqrt{2 \beta(t)} \sigma(t)\, d\omega_t}_{\text{Langevin diffusion component}},
    \end{align}
where $d\omega_t$ is the standard Wiener process. 
In principle, simulating either the Probability Flow ODE or the SDE above results in samples from the same distribution. 

\textbf{Training.} 
DM training reduces to learning a model $\vs_\vtheta(\rvx; \sigma)$ for the score function $\nabla_\rvx \log p(\rvx; \sigma)$. 
The model can, for example, be parameterized as $\nabla_\rvx \log p(\rvx; \sigma) \approx s_\vtheta(\rvx; \sigma) = (D_\vtheta(\rvx; \sigma) - \rvx)/ \sigma^2$~\citep{karras2022elucidating}, where $D_\vtheta$ is a learnable \emph{denoiser} that, given a noisy data point $\rvx_0 + \rvn$, $\rvx_0 \sim \pdata(\rvx_0)$, $\rvn \sim \gN\left(\bm{0}, \sigma^2 \mI_d\right)$, and conditioned on the noise level $\sigma$, tries to predict the clean $\rvx_0$. 
The denoiser $D_\vtheta$ (or equivalently the score model) can be trained via \emph{denoising score matching}~(DSM)
\begin{align} \label{eq:diffusion_objective}
    \E_{\substack{(\rvx_0, \rvc) \sim \pdata(\rvx_0, \rvc), (\sigma, \rvn) \sim p(\sigma, \rvn)}} \left[\lambda_\sigma \|D_\vtheta(\rvx_0 + \rvn; \sigma, \rvc) - \rvx_0 \|_2^2 \right],
\end{align}
where $p(\sigma, \rvn) = p(\sigma)\,\gN\left(\rvn; \bm{0}, \sigma^2\right)$, $p(\sigma)$ is a distribution over noise levels $\sigma$, $\lambda_\sigma \colon \R_+ \to \R_+$ is a weighting function, and $\rvc$ is an arbitrary conditioning signal, e.g., a class label, a text prompt, or a combination thereof. In this work, we choose $p(\sigma)$ to be a discrete distributions over 1000 noise levels and set $\lambda_\sigma = \sigma^{-2}$ similar to prior works~\cite{ho2020denoising,rombach2021high,sohl2015deep}.

\textbf{Classifier-free guidance.} Classifier-free guidance~\citep{ho2022classifier} is a technique to guide the iterative sampling process of a DM towards a conditioning signal $\rvc$ by mixing the predictions of a conditional and an unconditional model
\begin{align} \label{eq:guidance}
    D^w(\rvx; \sigma, \rvc) = (1 + w) D(\rvx; \sigma,\rvc) - w D(\rvx; \sigma),
\end{align}
where $w \geq 0$ is the \emph{guidance strength}. In practice, the unconditional model can be trained  jointly alongside the conditional model in a single network by randomly replacing the conditional signal $\rvc$ with a null embedding in~\Cref{eq:diffusion_objective}, e.g., 10\% of the time~\citep{ho2022classifier}. Classifier-free guidance is widely used to improve the sampling quality, trading for diversity, of text-to-image DMs~\citep{nichol2021glide, rombach2021high}.

\clearpage
\section{Comparison to the State of the Art}
\comparetoif

\newpage
\section{Comparison to Midjourney v5.1}

\subsection{Overall Votes}
To asses the generation quality of \modelname{} we perform a user study against the state of the art text-to-image generation platform Midjourney\footnote{We compare against v5.1 since that was the best version available at that time.}.
As the source for image captions we use the PartiPrompts (P2) benchmark~\cite{yu2022scaling}, that was introduced to compare large text-to-image model on various challenging prompts.

For our study, we choose five random prompts from each category, and generate four $1024 \times 1024$ images by both Midjourney (v5.1, with a set seed of 2) and \modelname{} for each prompt.  These images were then presented to the AWS GroundTruth taskforce, who voted based on adherence to the prompt. The results of these votes are illustrated in \Cref{fig:mjcomp_total}.
Overall, there is a slight preferance for \modelname{} over Midjourney in terms of prompt adherence.
\mjcompzero
\subsection{Category \& challenge comparisons on PartiPrompts (P2)}
Each prompt from the P2 benchmark is organized into a category and a challenge, each focus on different difficult aspects of the generation process.
We show the comparisons for each category (\Cref{fig:mjcomp_categories}) and challenge (\Cref{fig:mjcomp_challenges}) of P2 below.
In four out of six categories \modelname{} outperforms Midjourney, and in seven out of ten challenges there is no significant difference between both models or \modelname{} outperforms Midjourney.
\mjcompuno
\mjcompdue

\newpage
\section{On FID Assessment of Generative Text-Image Foundation Models}
\label{suppsec:fid_app}
\fidvsclip
Throughout the last years it has been common practice for generative text-to-image models to assess FID-~\cite{heusel2017gans} and CLIP-scores~\cite{radford2021learning,ramesh2021zeroshot} in a zero-shot setting on complex, small-scale text-image datasets of natural images such as COCO~\cite{lin2015microsoft}. %
However, with the advent of foundational text-to-image models~\cite{saharia2022photorealistic,ramesh2022hierarchical,rombach2021high,balaji2022ediffi}, which are not only targeting visual compositionality, but also at other difficult tasks such as deep text understanding, fine-grained distinction between unique artistic styles and especially a pronounced sense of visual aesthetics, this particular form of model evaluation has become more and more questionable. 
\citet{kirstain2023pick} demonstrates that COCO zero-shot FID is \emph{negatively correlated} with visual aesthetics, and such measuring the generative performance of such models should be rather done by human evaluators.
We investigate this for \modelname and visualize FID-vs-CLIP curves in \Cref{fig:fid+vs_clip} for 10k text-image pairs from COCO~\cite{lin2015microsoft}. Despite its drastically improved performance as measured quantitatively by asking human assessors (see \Cref{fig:userstudyandmodel}) as well as qualitatively (see \Cref{fig:comp_old_model} and \Cref{fig:comp_old_model_app}), \modelname does \emph{not} achieve better FID scores than the previous \sdshort versions. 
Contrarily, FID for \modelname is the worst of all three compared models while only showing slightly improved CLIP-scores (measured with OpenClip ViT~g-14). 
Thus, our results back the findings of~\citet{kirstain2023pick} and further emphasize the need for additional quantitative performance scores, specifically for text-to-image foundation models. All scores have been evaluated based on 10k generated examples.          

\newpage
\section{Additional Comparison between Single- and Two-Stage \modelname pipeline}
\refinevsnorefineappendix

\newpage
\section{Comparison between \textit{SD 1.5} vs. \textit{SD 2.1} vs. \modelname}
\sdoldcompapp

\sdoldcompapptwo
\FloatBarrier
\section{Multi-Aspect Training Hyperparameters}
\label{supsubsec:marlist}
We use the following image resolutions for mixed-aspect ratio finetuning as described in Sec.~\ref{subsec:mar}.

\begin{table}[ht]
\centering
\begin{minipage}{.45\textwidth}
\centering
\begin{tabular}{S[table-format=3] S[table-format=4] S}
\toprule
{\textbf{Height}} & {\textbf{Width}} & {\textbf{Aspect Ratio}} \\
\midrule
512 & 2048 & 0.25 \\
512 & 1984 & 0.26 \\
512 & 1920 & 0.27 \\
512 & 1856 & 0.28 \\
576 & 1792 & 0.32 \\
576 & 1728 & 0.33 \\
576 & 1664 & 0.35 \\
640 & 1600 & 0.4 \\
640 & 1536 & 0.42 \\
704 & 1472 & 0.48 \\
704 & 1408 & 0.5 \\
704 & 1344 & 0.52 \\
768 & 1344 & 0.57 \\
768 & 1280 & 0.6 \\
832 & 1216 & 0.68 \\
832 & 1152 & 0.72 \\
896 & 1152 & 0.78 \\
896 & 1088 & 0.82 \\
960 & 1088 & 0.88 \\
960 & 1024 & 0.94 \\
\bottomrule
\end{tabular}

\label{table:1}
\end{minipage}\hfill%
\begin{minipage}{.45\textwidth}
\centering
\begin{tabular}{S[table-format=4] S[table-format=4] S}
\toprule
{\textbf{Height}} & {\textbf{Width}} & {\textbf{Aspect Ratio}} \\
\midrule
1024 & 1024 & 1.0 \\
1024 & 960 & 1.07 \\
1088 & 960 & 1.13 \\
1088 & 896 & 1.21 \\
1152 & 896 & 1.29 \\
1152 & 832 & 1.38 \\
1216 & 832 & 1.46 \\
1280 & 768 & 1.67 \\
1344 & 768 & 1.75 \\
1408 & 704 & 2.0 \\
1472 & 704 & 2.09 \\
1536 & 640 & 2.4 \\
1600 & 640 & 2.5 \\
1664 & 576 & 2.89 \\
1728 & 576 & 3.0 \\
1792 & 576 & 3.11 \\
1856 & 512 & 3.62 \\
1920 & 512 & 3.75 \\
1984 & 512 & 3.88 \\
2048 & 512 & 4.0 \\
\bottomrule
\end{tabular}
\label{table:2}
\end{minipage}
\end{table}

\newpage
\section{Pseudo-code for Conditioning Concatenation along the Channel Axis}
\label{supsec:cond_pseudo_code}
\condcatcode

\clearpage
{\small
\bibliographystyle{plainnat}
\bibliography{arxiv, non_arxiv, postings}
}

\newpage

\end{document}